\documentclass[11pt]{article}

\newcommand{\toreview}[1]{{\color{black}#1}}

\newcommand{\revision}[1]{{\color{black}#1}}

\newcommand{\trace}{\textsc{Trace}}

\usepackage[preprint]{acl}

\usepackage{times}
\usepackage{latexsym}

\usepackage[T1]{fontenc}

\usepackage[utf8]{inputenc}

\usepackage{microtype}

\usepackage{inconsolata}

\usepackage{graphicx}

\usepackage{amsmath}
\usepackage{amssymb}
\usepackage{amsthm}
\usepackage{booktabs}
\usepackage{algorithm}
\usepackage{algorithmic}
\usepackage{multirow}
\usepackage{titletoc}
\usepackage{setspace}
\usepackage{subcaption}
\usepackage{xcolor}
\usepackage{float}
\usepackage[most]{tcolorbox}

\title{\trace: Business Rule-Grounded Reasoning Curriculum for Knowledge-Preserving Parametric Tool Retrieval in Enterprise LLMs}

\author{
        Sai Shruthi Sistla\thanks{Equal contribution.} \quad
        {\bf Ashutosh Hathidara}\footnotemark[1] \quad
        {\bf Christopher Toukmaji} \\
        {\bf Mayank Shrivastava} \quad
        {\bf Karthikeyan Asokkumar} \\[6pt]
        SAP Labs \\
        {\small \texttt{\{sai.shruthi.sistla, ashutosh.hathidara, chris.toukmaji,}} \\
        {\small \texttt{mayank.shrivastava02, karthikeyan.asokkumar\}@sap.com}}
      }

\begin{document}
\maketitle
\begin{abstract}
\toreview{Parametric retrieval enables LLMs to retrieve tools implicitly by assigning each API a unique virtual token and training the model to generate it via constrained beam search~\citep{wang2025toolgen}. Toolsense~\citep{toolsense2026} show that this regime has two critical drawbacks: it destroys parametric tool knowledge during training, and its beam-search decoding is too slow for real-time deployment. We introduce \trace{} (\textbf{T}ool \textbf{R}etrieval via \textbf{A}ugmented \textbf{C}hain-of-thought and \textbf{E}nterprise rules), a two-stage curriculum that resolves this dissociation. Stage~1 reuses the multi-format memorization SFT from ToolSense to seed tool knowledge with LoRA. Stage~2 is our core contribution: the model is trained to emit a thinking trace before producing a JSON list of tool tokens, using two data sources --- RRB pairs from ToolSense and queries synthesized to target business rules curated by domain experts --- both augmented with reasoning traces. This training objective preserves Stage~1 MCQ and QA probing accuracy while enabling single-beam greedy decoding at production latency. Evaluated on a combined enterprise catalog of \textbf{8,283} tools across two enterprise product lines, \trace{} training for Stage~2 not only preserves but \textit{improves} tool understanding: MCQ accuracy gains \textbf{+3.2\,pp} and QA probing gains \textbf{+9\,pp} over Stage~1. On retrieval, \trace{} achieves \textbf{$\sim$86\%} recall on Domain~A and \textbf{$\sim$60\%} on Domain~B --- compared to embedding baseline performance of $\sim$27\% \& $\sim$52\% --- both with single-beam greedy decoding, making it directly deployable at production latency.}
\end{abstract}

\section{Introduction}
\label{sec:intro}

\toreview{Routing user queries to the right API across thousands of proprietary enterprise tools is a core challenge for AI copilots. Embedding-based retrieval~\citep{karpukhin2020dpr} treats this as semantic matching, but plateaus at low recall on terse, overlapping tool descriptions (\S\ref{subsec:retrieval}) and \emph{critically} never internalizes tool knowledge, foreclosing richer downstream capabilities. Parametric tool retrieval~\citep{wang2025toolgen} in LLMs offers solution to both problems, but its retrieval training objective catastrophically destroys parametric tool knowledge in the process~\citep{toolsense2026}.}

\toreview{In our production enterprise AI copilot serving $\sim$\textbf{39k} monthly active users, embedding-based retrieval accounts for \textbf{$\sim$60\%} of incorrect tool retrievals, the dominant failure mode at scale and a hard bottleneck for the entire system, since no downstream component can recover from retrieving the wrong tool. Analysis of 609 stratified sampled sessions reveals two recurring patterns: \textbf{$\sim$81.7\%} of failures involve many ($>10$) semantically overlapping tools that require domain business rules to disambiguate --- API deprecations, versioning changes, and domain-specific routing constraints --- that embedding models cannot track, leaving both the retriever and any downstream model blind to the rules that govern correct tool selection. These failure modes cannot be fixed by improving the embedding model alone; they demand a retrieval system that can efficiently \emph{reason} about tools.}

\begin{figure*}[h]
    \centering
    \includegraphics[width=\linewidth, trim = 0.5cm 0cm 0.5cm 0cm, clip]{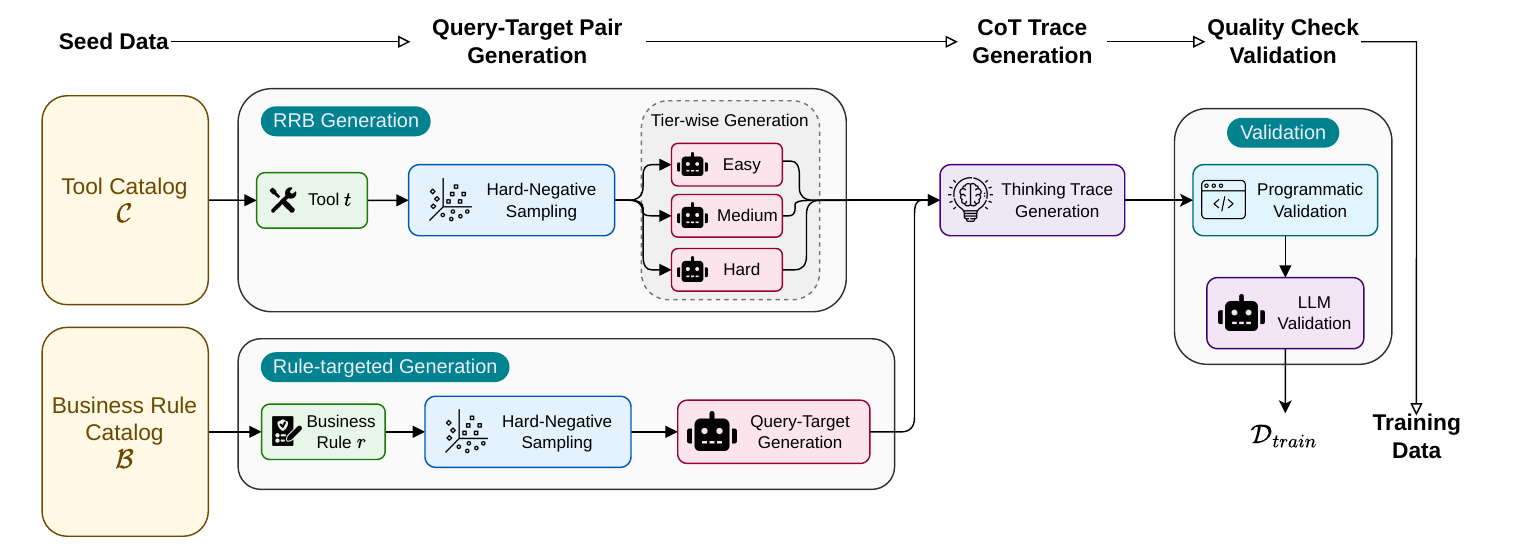}
    \caption{Methodology for Generating Data $\mathcal{D}_{train}$ for Stage~2 \trace{} training}
    \label{fig:trace_generation_pipeline}
\end{figure*}

\toreview{ToolGen~\citep{wang2025toolgen} introduced a paradigm to enable LLMs to retrieve tools implicitly by assigning each tool a unique virtual token string (e.g. \texttt{<<WeatherAPI/GetForecast>>}) and training the model to generate that token given a user query. Constrained beam search over a prefix trie of valid token strings enables the tool retrieval, given a query. Later, ToolSense~\citep{toolsense2026} performed empirical analysis also on tool representation made of hierarchical tokens (e.g. \texttt{<<WeatherAPI>><<GetForecast>>}). ToolGen achieves a high recall of more than 90\% on in-distribution benchmark for the ToolBench~\citep{qin2024toolllm} catalog of $\sim$47,000+ tools.}

\toreview{ToolSense ~\citep{toolsense2026} introduced method to synthesize 3 benchmarks: Realistic Retrieval Benchmark (RRB) to evaluate retrieval performance on out-of-distribution queries, MCQ probing benchmark to assess multiple-choice, and QA probing benchmark to evaluate factual yes/no question answering. Empirical analysis on the same ToolBench catalog revealed that the model capable of high retrieval recall on the in-distribution evaluation benchmark doesn't generalize to the out-of-distribution RRB benchmark, and more importantly, the retrieval training to map queries to token strings catastrophically destroys the model's parametric tool knowledge as measured by MCQ and QA probing. This dissociation between retrieval performance and parametric tool knowledge raises a critical question: \emph{can we design a training curriculum that preserves tool knowledge while enabling retrieval?}}

\toreview{We introduce \trace{}, a two-staged curriculum that resolves the knowledge-retrieval dissociation to enable faithful retrieval with reasoning at production scale. LLM is trained to reason-and-retrieve using data synthesized with business grounded \& ungrounded query-reasoning pairs, and the inference is enabled at single beam greedy decoding that helps reduce latency to meet production requirements. On an 8,283-tool enterprise catalog, \trace{} achieves \textbf{$\sim73\%$} retrieval recall on combined tools from two domains while preserving parametric tool knowledge (\textbf{+28\,pp} MCQ$_\text{expert}$ over the non-reasoning baseline), with \textbf{${\sim}200{\times}$} higher throughput than constrained beam search.}

\toreview{To summarize, our contributions are threefold: (1)~\textbf{\trace{} curriculum}: a two-stage training recipe that preserves tool knowledge while delivering competitive retrieval performance with single-beam greedy decoding at production latency; (2)~\textbf{Business rule grounding}: a data augmentation mechanism that injects enterprise business rules into reasoning traces, lifting retrieval on semantically overlapping tools by up to $+$23~pp; (3)~\textbf{Production deployment study}: end-to-end validation on a combined catalog of 8,283 proprietary enterprise tools across two business domains, establishing reasoning retrieval as the knowledge-safe path for enterprise AI copilots.} \revision{Beyond retrieval, the preserved tool knowledge provides a foundation for downstream agentic tool planning without external retrieval dependencies.}

\section{Related Work}
\label{sec:related}

\paragraph{Tool retrieval and function calling.}
\toreview{A large body of work trains LLMs to use external tools, from function-calling fine-tuning \citep{schick2023toolformer,tang2023toolalpacageneralizedtoollearning} to full API-call generation \citep{patil2024gorilla,qin2024toolllm}. These approaches assume the correct tool is provided in context or drawn from a small candidate set; they do not address \emph{tool retrieval}: identifying the right tool from a large catalog where the answer is unknown, a distinct problem in which catalog size defeats in-context enumeration and semantic overlap makes embedding matching unreliable. \citet{wang2025toolgen} closes this gap with ToolGen, mapping each tool to a unique virtual token generated via constrained beam search over a prefix trie. ToolSense~\citep{toolsense2026} shows that such training systematically destroys learned tool knowledge, foreclosing downstream post-training that relies on it. \trace{} treats tool-knowledge preservation as a first-class training objective while eliminating the external retriever at inference time.}


\paragraph{Reasoning traces in supervised fine-tuning.}
\toreview{Chain-of-thought prompting \citep{3600270.3602070} showed that eliciting intermediate reasoning improves LLM performance, and subsequent work distils this via fine-tuning: STaR~\citep{zelikman2022star}, Orca~\citep{mukherjee2023orcaprogressivelearningcomplex}, and DeepSeek-R1~\citep{Guo_2025}. These target generic reasoning capabilities, whereas \trace{} applies reasoning SFT to parametric retrieval in order to \emph{preserve} the tool knowledge that the non-reasoning retrieval objective destroys.}


\paragraph{Catastrophic forgetting and knowledge retention.}
\toreview{Narrow-objective fine-tuning is known to degrade acquired knowledge \citep{11151751}, and \citet{biderman2024lora} show LoRA forgets less than full fine-tuning, grounding Stage~1 design. \citet{toolsense2026} and \citet{mallen-etal-2023-trust} establish probing as a reliable measure of parametric factual knowledge, which we adapt to the enterprise tool setting: LoRA at Stage~1 buffers retention across stages, and MCQ/QA probing is our primary diagnostic for the knowledge-retrieval dissociation.}


\section{\trace: Two-Stage Curriculum with Reasoning Traces}
\label{sec:trace}

\toreview{\trace{} trains an LLM to \emph{reason-then-retrieve}: given a user query, the model first emits a thinking trace that deliberates over candidate tools in its parametric memory and the business rules that govern them, then commits to a JSON list of virtual tool tokens. Training follows a two-stage curriculum. \textbf{Stage~1} (\S\ref{subsec:stage1}) seeds parametric tool knowledge via multi-format memorization SFT with LoRA, reusing the recipe of ToolSense~\citep{toolsense2026}. \textbf{Stage~2} (\S\ref{subsec:stage2-data}) is our core contribution: a reasoning-augmented retrieval SFT trained on a corpus $\mathcal{D}_{train}$ synthesized by the three-phase pipeline in Figure~\ref{fig:trace_generation_pipeline} --- query-target pair generation, CoT trace generation, and quality-check validation. We first formalize the setting and seed data (\S\ref{subsec:setup}) before describing each stage.}

\subsection{Problem Setup and Seed Data}
\label{subsec:setup}
 
\toreview{We are given an enterprise tool catalog of $|N| = 8{,}283$ tools spanning two domains: Domain~A (HR; $918$ tools) and Domain~B (Finance; $7{,}365$ tools), $\mathcal{C} = \big\{\, t_i = (\mathrm{name}_i, \mathrm{desc}_i) \,\big\}_{i=1}^{|N|}$, where each tool $t$ is assigned a unique virtual token $v_t$ appended to the model vocabulary, following the parametric retrieval paradigm~\citep{wang2025toolgen}. In addition and unique to the enterprise setting, we are given a business rule catalog of size $|M| = 123$, $\mathcal{B} = \big\{\, r_j = \big(\mathcal{T}(r_j),\, \mathrm{text}(r_j)\big) \,\big\}_{j=1}^{|M|}$ curated by domain experts. Each rule $r_j$ is a pair consisting of (i)~a \emph{confusible tool set} $\mathcal{T}(r_j) \subseteq \mathcal{C}$, a small group of tools whose surface descriptions overlap heavily and differ only in fine-grained, domain-specific details, and (ii)~a natural-language rule body $\mathrm{text}(r_j)$ that specifies how to disambiguate among the tools in $\mathcal{T}(r_j)$, capturing API deprecations, versioning changes, and domain-specific routing constraints (e.g., ``for product line~X, requests about year 2024 and later must route to API ABC and earlier should route to API PQR.''). These rules are precisely the knowledge that is only limited to domain-experts and domain-developers, and is not inferable from tool descriptions alone, making them critical for correct retrieval.}
 
\toreview{Given a query $q$, the retrieval task produces the tool set $\hat{A}(q) \subseteq \mathcal{C}$. \trace{} models this as conditional generation of a trace--answer pair $(\hat{z}, \hat{y})$, where $\hat{z}$ is a free-form thinking trace and $\hat{y}$ is an ordered list of virtual tool tokens $[\, v_x : x \in \hat{A}(q) \,]$. Later, this inference result is compared with the corresponding ground truth $y$ and $A(q)$.}

\subsection{Stage 1: Multi-Format Memorization}
\label{subsec:stage1}
 
\toreview{Stage~1 seeds parametric tool knowledge by fine-tuning the base LLM on multiple complementary views of every tool $t \in \mathcal{C}$, following the multi-format memorization recipe of ToolSense~\citep{toolsense2026} (details in Appendix~\ref{app:multi-format-memo}). The objective is not retrieval but \emph{internalization}: by Stage~2, every $v_t$ must be internalized in the model's representation space, tightly bound to the tool's semantics, name, and surface form.}

\subsection{Stage 2: Reasoning-Augmented Retrieval}
\label{subsec:stage2-data}

\toreview{Stage~2 trains the model to emit a thinking trace $\hat{z}$ before committing to an ordered list of virtual-tokens $\hat{y}$. Its training corpus $\mathcal{D}_{\mathrm{train}}$ is synthesized by the three-phase pipeline of Figure~\ref{fig:trace_generation_pipeline}; we summarize the design choices that drive performance and defer mechanics to Appendix~\ref{app:stage2-pipeline}.}

\toreview{\textbf{Two-branch query generation.} We synthesize $(q, A)$ pairs from two complementary branches. \emph{(i)~RRB generation} instantiates the ToolSense RRB pipeline~\citep{toolsense2026} over $\mathcal{C}$: for each anchor tool we form a hard-negative pool from its top-$K$ nearest neighbours under a sentence encoder, and prompt a generator LLM at three difficulty tiers ($|A|{=}1$, $|A|\!\in\!\{2,3\}$, $|A|\!\geq\!4$) to elicit the ambiguity spectrum of real user queries. \emph{(ii)~Rule-targeted generation} iterates over every $r \in \mathcal{B}$ and builds its pool from $\mathcal{T}(r)$ together with each confusible's nearest neighbours. For each rule we synthesize queries in three phrasing styles --- \emph{explicit} (keywords directly signal the rule), \emph{implicit} (the rule applies but must be inferred from context without obvious signals), and \emph{exception} (the query appears to match the rule's primary tool but an edge case redirects to a confusable). Such rule-targeted generation densifies coverage of domain understanding through user-like queries that require business rules to disambiguate between confusables.}

\toreview{\textbf{Trace generation: name-token coupling.} A teacher LLM produces a reference trace $z$ following a fixed deliberative schema: scope the user's access permissions, identify and compare relevant APIs, narrow to endpoint candidates, apply the governing business rule (quoting it by name), and commit to the final tool selection. Both in reference $z$ and the answer reference tools, the tool names at training time are replaced by its virtual token $v_t$, yielding the supervision target $o= (\tilde{z},\, y)$ where $\tilde{z}$ is the name-substituted trace and $y = \mathrm{JSON}([\, v_a : a \in A \,])$. This substitution is the mechanism by which Stage~2 \emph{reinforces} Stage~1 knowledge: every $v_a$ appears inside the deliberative context that justifies it (``\ldots the billing document domain is handled by $v_{t_1}$, but the rule states header-level queries belong to $v_{t_2}$,\ldots''), preventing $v_a$ from decoupling from $\mathrm{desc}_a$.}

\toreview{\textbf{Validation.} Each candidate $(q, z, A)$ passes a two-filter cascade: a deterministic filter enforcing grounding, no-leakage, JSON consistency, and rule attribution; followed by an LLM judge scoring naturalness, label correctness, and \emph{trace faithfulness}. Failures are regenerated with judge feedback up to a retry budget of 5; samples that exhaust retries are dropped. Surviving samples form $\mathcal{D}_{\mathrm{train}} = \{(q_i, z_i, A_i)\}_{i=1}^{D}$. Filter definitions, judge rubric, and yield statistics are in Appendix~\ref{app:stage2-pipeline}.}

\toreview{Both stages use standard autoregressive SFT with prompt-side loss masking, following the ToolSense settings (Appendix \ref{app:training}). At inference, \trace{} decodes greedily in a single beam --- emitting the trace $\hat{z}$ followed by the JSON token list $\hat{y}$ --- with no constrained beam search or prefix-trie expansion, making it directly deployable at production latency (Appendix~\ref{app:inference}).}

\section{Experiments and Results}
\label{sec:experiments}

\toreview{We evaluate \trace{} on a combined enterprise catalog of 8,283 tools and our experiments answer five questions: (1)~Does constrained retrieval training destroy tool knowledge as measured by ToolSense? (2)~Does \trace{}'s curriculum resolve this dissociation? (3)~How much does business-rule grounding help? (4)~Does the recipe generalize across token formats? (5)~Does this recipe deliver production-latency with competitive retrieval and preserved knowledge?}

\subsection{Experimental Setup}
\label{subsec:setup-exp}

\toreview{All experiments use \textbf{Gemma4-E4B-it} ~\citep{googledeepmind2026gemma4e4bit} as the base model (Appendix \ref{app:training}) with combined tool catalog and metrics are reported per-domain to surface domain-specific behavior.}

\toreview{\textbf{Evaluation suite.} We evaluate retrieval on \textbf{PRB} (Production Retrieval Benchmark), a hand-curated set of golden production query-tool pairs from our enterprise copilot ($n_A{=}131$, $n_B{=}123$). \revision{PRB queries have no semantic overlap with the synthetic training data; no PRB query or paraphrase was used as input to training data generator.} We report recall ($R$) under two decoding views: constrained beam search over the top-10 candidates per query (\textit{R@10}), and greedy single-beam reasoning where the model commits to its own answer set $\hat{A}(q)$ (\textit{R@gen}). We probe parametric tool knowledge with three benchmarks: \textbf{MCQ$_\text{ts}$} and \textbf{QA$_\text{ts}$}, multiple-choice and yes/no benchmarks synthesized from $\mathcal{C}$ via the ToolSense ~\citep{toolsense2026} ($300{+}400$ and $200{+}400$ questions across the two domains; random baselines $25\%$ and $50\%$); and \textbf{MCQ$_\text{expert}$}, a $454$-question, expert-authored multiple-choice benchmark held out from training (random baseline $29.3\%$ from non-uniform answer-set sizes; details in Appendix~\ref{app_subsec:eval-set-construction-mcq-expert}). MCQ$_\text{expert}$ is our primary knowledge diagnostic; the synthesized probes serve as in-distribution controls.}


\toreview{\textbf{Configuration axes.} Every run is described by a triple $(F, M, R)$ enumerated in Table~\ref{tab:config-grid}: token format $F$, Stage~1 memorization recipe $M$, and Stage~2 retrieval recipe $R$. The three formats are \texttt{a}~flat (e.g.\ \texttt{<<GetForecast>>}), \texttt{b}~bare-hierarchical (e.g.\ \texttt{<<WeatherAPI>><<GetForecast>>}), and \texttt{c}~wrapped-hierarchical (e.g.\ \texttt{<tid><<API>><<EP>></tid>}).}

\begin{table}[t]
\centering
\small
\begin{tabular}{lp{5.0cm}}
\toprule
Axis & Levels \\
\midrule
\textbf{F} (format) & \texttt{a}~flat, \texttt{b}~bare-hier, \texttt{c}~wrapped-hier \\
\textbf{M} (Stage~1) & $\varnothing$~none, \texttt{m}~memo, \texttt{mr}~reasoning memo \\
\textbf{R} (Stage~2) & $\varnothing$~none, \texttt{n}~non-reasoning, \texttt{r}~reasoning, \texttt{r+R}~reasoning + rules \\
\bottomrule
\end{tabular}
\caption{Configuration grid.}
\label{tab:config-grid}
\end{table}

\subsection{Knowledge Preservation}
\label{subsec:preservation}

\begin{figure}[t]
  \centering
  \includegraphics[width=0.9\columnwidth, trim = 0.5cm 0.3cm 0cm 0cm, clip]{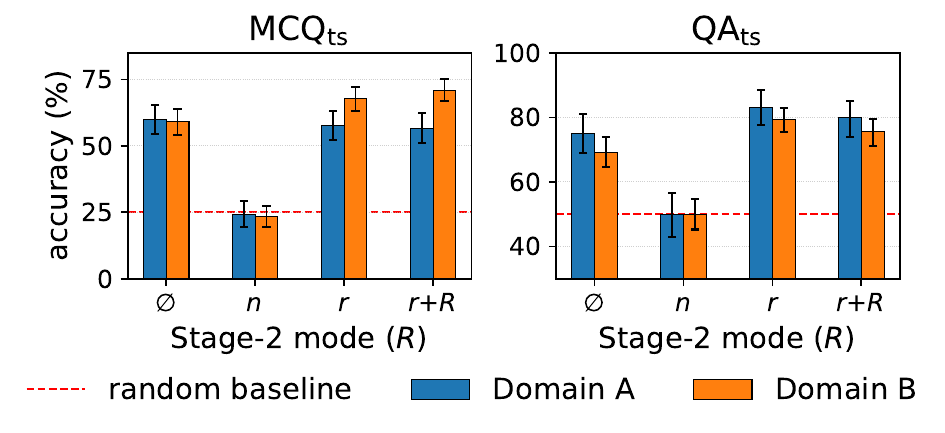}
  \caption{\textbf{In-distribution probing.} MCQ$_\text{ts}$ (left) and
  QA$_\text{ts}$ (right) accuracy across the four Stage-2 modes.}
  \label{fig:knowledge}
\end{figure}

\toreview{Figure~\ref{fig:knowledge} reports MCQ$_\text{ts}$ and QA$_\text{ts}$ for the four Stage-2 modes on the wrapped-hierarchical format ($F{=}c$). Full per-format results are in Appendix~\ref{app_subsec:probing_evaluations}.}

\begin{figure}[t]
  \centering
  \begin{subfigure}{0.48\columnwidth}
    \includegraphics[width=\linewidth]{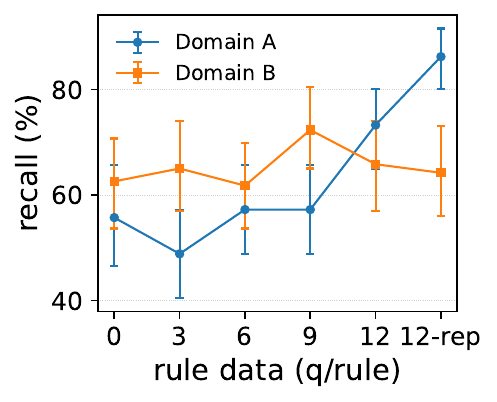}
    \caption{Retrieval recall ($R@gen$) on the two domains.}
    \label{fig:pareto-recall}
  \end{subfigure}
  \hfill
  \begin{subfigure}{0.48\columnwidth}
    \includegraphics[width=\linewidth]{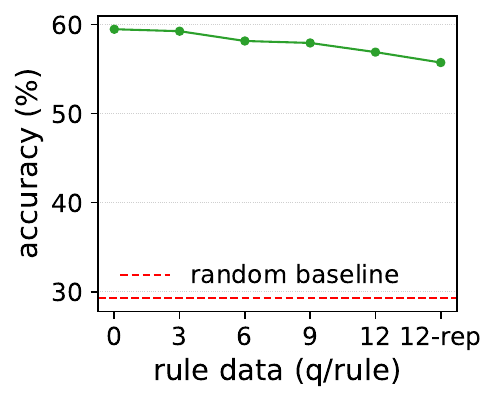}
    \caption{MCQ$_\text{expert}$ accuracy (random baseline $29.3\%$).}
    \label{fig:pareto-heldout}
  \end{subfigure}
  \caption{\textbf{Stage~2 Rule-grounded Pareto sweep} on $F{=}\textsf{a}$,
  $M{=}\textsf{m}$ and $R=r+R$ where we vary the number of queries per rule.}
  \label{fig:pareto}
\end{figure}

\toreview{The pattern is consistent across all three probes. Non-reasoning retrieval training ($R{=}n$, the ToolGen recipe) collapses every benchmark to near its random baseline, replicating the dissociation finding of~\citep{toolsense2026} on a fundamentally different enterprise catalog and on a held-out probe authored by domain experts. Reasoning retrieval ($R{=}r$, $r{+}R$) reverses the collapse and even improves on MCQ$_\text{ts}$ and QA$_\text{ts}$ over the Stage-1 ceiling. The same pattern carries over to the MCQ$_\text{expert}$ probe (random baseline $29.3\%$): {\small$\,\varnothing{:}\,69.2\;\;n{:}\,33.7\;\;r{:}\,61.5\;\;r{+}R{:}\,56.4\,$}.} \revision{The gap between $R{=}r$ and $R{=}r{+}R$ is a \emph{navigable trade-off}: rule grounding costs only $5.1$\,pp on MCQ$_\text{expert}$ (vs the $35.5$\,pp collapse under non-reasoning) in exchange for a $+32$\,pp retrieval gain on Domain~A (Figure~\ref{fig:pareto}). The in-distribution tool probes (MCQ$_\text{ts}$, QA$_\text{ts}$) stay within their confidence intervals across the entire rule-data axis (Table~\ref{tab:app-rule-scaling}), so rule grounding does not erode tool-specific knowledge; the residual cost reflects production-correct routing logic that diverge from generic domain interpretation.}

\subsection{Retrieval Performance}
\label{subsec:retrieval}

\toreview{We compare \trace{} against the non-reasoning retrieval baseline $(c, m, n)$ on the wrapped-hierarchical format under both decoding views, and then map the \emph{rule-grounded data} axis to expose the Pareto trade-off between retrieval recall and expert tool knowledge.}

\begin{table}[t]
  \centering\small
  \begin{tabular}{lcccc}
    \toprule
    & \multicolumn{2}{c}{Domain~A} & \multicolumn{2}{c}{Domain~B} \\
    \cmidrule(lr){2-3}\cmidrule(lr){4-5}
    Method ($F, M, R$) & R@1 & R@10 & R@1 & R@10 \\
    \midrule
    text-embedding-3-large & 13.74 & 27.48 & 27.59 & 52.71 \\
    ToolGen $(c, m, n)$           & 45.0 & \textbf{87.0} & 35.8 & 78.0 \\
    \trace$_{-R}$ $(c, m, r)$     & 40.5 & 74.0 & \textbf{46.3} & \textbf{79.7} \\
    \trace{} $(c, m, r{+}R)$      & \textbf{48.9} & 86.3 & 36.6 & 77.2 \\
    \bottomrule
  \end{tabular}
  \caption{Fair retrieval comparison between non-reasoning \& reasoning variants under 10-beam constrained decoded inference. $F{=}\textsf{c}$ for all variants.}
  \label{tab:headhead-vb}
\end{table}

\begin{table}[t]
  \centering\small
  \begin{tabular}{lcccc}
    \toprule
    & \multicolumn{2}{c}{Domain~A} & \multicolumn{2}{c}{Domain~B} \\
    \cmidrule(lr){2-3}\cmidrule(lr){4-5}
    Method ($F, M, R$) & R@1 & R@gen & R@1 & R@gen \\
    \midrule
    \trace$_{-R}$ $(c, m, r)$     & 32.1 & 53.4 & \textbf{56.1} & \textbf{67.5} \\
    \trace{} $(c, m, r{+}R)$      & \textbf{51.1} & \textbf{85.5} & 43.1 & 60.2 \\
    \bottomrule
  \end{tabular}
  \caption{Comparison of reasoning retrieval variants under no constrained decoding and single beam with JSON list of tool tokens as output. $F{=}\textsf{c}$ for all variants.}
  \label{tab:headhead-va}
\end{table}

\toreview{\textbf{Comparison with non-reasoning retrieval.} Tables~\ref{tab:headhead-vb} \& ~\ref{tab:headhead-va} report PRB recall for the three Stage-2 variants on $F{=}\textsf{c}$. The embedding baseline (\texttt{text-embedding-3-large}) reaches only $27.5\%$/$52.7\%$ R@10 on Domain~A/B, confirming semantic matching alone is insufficient. Under constrained decoding, \trace{} with rule grounding $(c, m, r{+}R)$ matches the non-reasoning baseline on R@10 and slightly improves R@1 on Domain~A; without rule grounding, $(c, m, r)$ trails by $\sim\!13$\,pp R@10, so rule data is necessary for retrieval-grade recall, not a free add-on. Under single-beam reasoning decoding, $(c, m, r{+}R)$ retains $85.5\%$ R@gen on Domain~A while Domain~B sits at $60.2\%$. This gap reflects \emph{incomplete rule coverage}: $\mathcal{B}$ does not yet cover every confusable cluster in the $7{,}365$-tool Domain~B catalog, so rule-grounded queries help the covered subset at the expense of the rest. This is a coverage artifact, not a method limitation: rules do not regress Domain~B on $F{=}\textsf{a}$ ($62.6\!\to\!64.2$). Full results appear in Appendix~\ref{app_subsec:retrieval_evals}.}


\revision{\textbf{Comparison with hybrid retrieval.} We additionally compare against a strong hybrid pipeline: \texttt{text-embedding-3-large} retrieves the top-50 candidates, followed by an LLM reranker, \texttt{claude-4.6-sonnet}~\citep{anthropic2026claudesonnet46} (with and without business rules injected at inference time). \trace{} outperforms both hybrid variants on R@gen: $85.5\%$ vs.\ $67.2\%$ (no rules) \& $66.4\%$ (with rules) on Domain~A, and $60.2\%$ vs.\ $54.5\%$ \& $56.1\%$ on Domain~B, demonstrating that parametric internalization of rules during training is more effective than inference-time rule injection (Appendix~\ref{app:hybrid-baseline}).}


\subsection{Business Rule Grounding}
\label{subsec:rule-grounding}

\toreview{Figure~\ref{fig:pareto} sweeps rule-data density (0--12 query/rule, plus a replace strategy that swaps RRB pairs for rule-targeted ones) on $F{=}\textsf{a}$, measuring PRB recall and MCQ$_\text{expert}$. All-format results in Appendix~\ref{app_subsec:retrieval_evals}.}

\toreview{\textbf{Retrieval scaling.} On Domain~A, R@gen climbs from $55.7\%$ at $0$~q/rule to $73.3\%$ at $12$ ($+17.6$~pp) and to $\mathbf{86.3\%}$ under the replace strategy ($+30.6$~pp). Domain~B, where the rule catalog is still sparse (\S\ref{subsec:retrieval}), stays within its baseline CI across the sweep --- the gain manifests only where rules cover the confusable clusters.}

\toreview{\textbf{Knowledge cost.} MCQ$_\text{expert}$ decreases monotonically from $59.5\%$ at $0$~q/rule to $55.7\%$ at $12$-\textit{rep} (a $3.8$~pp drop). The decline is modest given the probe targets general domain knowledge rather than tool knowledge specifically, and the in-distribution probes MCQ$_\text{ts}$ and QA$_\text{ts}$ stay within their CIs across the same axis (Table~\ref{tab:app-rule-scaling}).}

\toreview{\textbf{Rule citation analysis.} To confirm the model genuinely reasons about rules rather than memorizing answer mappings, we analyze the generated traces of $(c, m, r{+}R(12{-}rep))$ on Domain~A PRB. After rule-grounded training, the rule citation rate rises from $0\%$ (no-rules baseline) to $71\%$ of traces; on traces that cite a rule, recall reaches $94.6\%$, versus $63.2\%$ on traces that do not.} \revision{A human labelling study on both PRB benchmarks further validates this: when the ground-truth query requires a business rule, the model correctly cites the relevant rule $82.4\%$/$79.5\%$ of the time (Domain~A/B), achieving $71.2\%$/$75.3\%$ correct retrieval on those samples, providing converging evidence of improved retrieval due to rule inference (Appendix~\ref{app:human-study}).}

\subsection{Token Format Generalization}
\label{subsec:token-format}

\toreview{Repeating the pipeline across all three token formats from \S\ref{subsec:setup-exp} confirms the dissociation and its resolution are not format-specific. Non-reasoning retrieval ($R{=}\textsf{n}$) collapses MCQ$_\text{expert}$ to near random in every format ($34.8\%$, $30.0\%$, $33.7\%$ for $F\in\{\textsf{a},\textsf{b},\textsf{c}\}$); reasoning retrieval ($R{=}\textsf{r}$) recovers it to $59.5$, $50.9$, $61.5$. PRB R@gen tracks the same direction (Table~\ref{tab:app-format-sweep}): $F{=}\textsf{a}$ and $F{=}\textsf{c}$ both reach $\sim\!86\%$ on Domain~A under the rule-grounded configuration, while $F{=}\textsf{b}$ lags by $\sim\!10$\,pp on every metric.}

\toreview{The $\textsf{n}{\to}\textsf{r}$ curriculum (bolting reasoning onto a non-reasoning checkpoint) shows a sharper format dependence: on $F{=}\textsf{a}$ it partially recovers MCQ$_\text{expert}$ to $49.8\%$, but on $F{=}\textsf{b}$ and $F{=}\textsf{c}$ it remains at the random-baseline floor ($28.2\%$ and $31.5\%$). The $\textsf{n}{\to}\textsf{r}$ result is not part of the headline curriculum, but it strengthens the case for avoiding the non-reasoning retrieval objective altogether: damage caused by $R{=}\textsf{n}$ is, at best, only partly reversible.}

\subsection{Latency Improvement}
\label{subsec:latency}


\toreview{We compare end-to-end inference latency on the same \trace{} headline checkpoint $(c, m, r{+}R(12{-}rep))$ under (i)~single-beam free-form decoding with reasoning and (ii)~10-beam constrained decoding without reasoning, both served by vLLM~\citep{10.1145/3600006.3613165} on a single H200 with $100$ Domain-A PRB queries. free-form decoding answers a single user in $\sim\!1.9$~s versus $\sim\!19$~s for constrained beam-10, and at concurrency $32$ it sustains $\mathbf{11.2}$~qps versus $\mathbf{0.05}$~qps, a $\sim\!200{\times}$ throughput gap (Figure~\ref{fig:latency} in Appendix~\ref{app:inference}).}

\section{Conclusion}
\label{sec:conclusion}

 We presented \trace{}, a two-stage curriculum for parametric tool retrieval that decouples tool knowledge from the retrieval objective via a business-rule-grounded reasoning trace. Stage~1 memorizes the tool catalog; Stage~2 trains the model to reason over business rules before emitting tool tokens. On 8{,}283 enterprise tools across two domains, \trace{} reaches $\sim$86\% recall on Domain~A and $\sim$60\% on Domain~B under single-beam greedy decoding, while improving Stage-1 MCQ and QA probing by +7.6\,pp and +4.5\,pp respectively. \trace{} thus delivers parametric tool retrieval that is knowledge-preserving and deployable at production latency.

\section*{Limitations}
\label{sec:limitations}

\trace{} is evaluated within a deliberately scoped regime for enterprise use case, and three boundaries of that regime are (i) our experiments cover two enterprise domains (HR and Finance) at a single model size, and behavior at higher parameters size or in adjacent verticals (e.g., procurement, healthcare) is an open question. (ii) the benchmarks corresponds to proprietary tool catalogs and thus cannot be released, though we describe the full details in appendix so that the protocol can be reproduced on any analogous catalog. (iii) the business-rule corpus underpinning Stage~2 is authored by domain experts --- which is exactly what gives the rules their grounding power, but also means the curation cost is non-trivial; we do not evaluate automatic rule extraction, and identifying when a rule corpus is sufficiently dense (\S\ref{subsec:retrieval} suggests $\geq$6 queries per rule) remains a recipe rather than a closed-form criterion. Finally, on the larger Domain~B catalog (7,365 tools) greedy decoding still trails constrained-beam; closing this scale gap --- through richer rule coverage --- is the most concrete future direction.

\section*{Ethical Considerations}
\label{sec:ethics}

We conducted experiments within the provisions of the ACL Ethics Policy and relevant research-integrity guidelines. There are, to the best of our knowledge, no remaining ethical risks that have not been addressed.

\bibliography{custom}

@inproceedings{
    wang2025toolgen,
    title={ToolGen: Unified Tool Retrieval and Calling via Generation},
    author={Renxi Wang and Xudong Han and Lei Ji and Shu Wang and Timothy Baldwin and Haonan Li},
    booktitle={The Thirteenth International Conference on Learning Representations},
    year={2025},
    url={https://openreview.net/forum?id=XLMAMmowdY}
}

@inproceedings{
    qin2024toolllm,
    title={Tool{LLM}: Facilitating Large Language Models to Master 16000+ Real-world {API}s},
    author={Yujia Qin and Shihao Liang and Yining Ye and Kunlun Zhu and Lan Yan and Yaxi Lu and Yankai Lin and Xin Cong and Xiangru Tang and Bill Qian and Sihan Zhao and Lauren Hong and Runchu Tian and Ruobing Xie and Jie Zhou and Mark Gerstein and dahai li and Zhiyuan Liu and Maosong Sun},
    booktitle={The Twelfth International Conference on Learning Representations},
    year={2024},
    url={https://openreview.net/forum?id=dHng2O0Jjr}
}

@inproceedings{
    patil2024gorilla,
    title={Gorilla: Large Language Model Connected with Massive {API}s},
    author={Shishir G Patil and Tianjun Zhang and Xin Wang and Joseph E. Gonzalez},
    booktitle={The Thirty-eighth Annual Conference on Neural Information Processing Systems},
    year={2024},
    url={https://openreview.net/forum?id=tBRNC6YemY}
}

@inproceedings{
    schick2023toolformer,
    title={Toolformer: Language Models Can Teach Themselves to Use Tools},
    author={Timo Schick and Jane Dwivedi-Yu and Roberto Dessi and Roberta Raileanu and Maria Lomeli and Eric Hambro and Luke Zettlemoyer and Nicola Cancedda and Thomas Scialom},
    booktitle={Thirty-seventh Conference on Neural Information Processing Systems},
    year={2023},
    url={https://openreview.net/forum?id=Yacmpz84TH}
}

@misc{tang2023toolalpacageneralizedtoollearning,
      title={ToolAlpaca: Generalized Tool Learning for Language Models with 3000 Simulated Cases}, 
      author={Qiaoyu Tang and Ziliang Deng and Hongyu Lin and Xianpei Han and Qiao Liang and Boxi Cao and Le Sun},
      year={2023},
      eprint={2306.05301},
      archivePrefix={arXiv},
      primaryClass={cs.CL},
      url={https://arxiv.org/abs/2306.05301}, 
}

@inproceedings{3600270.3602070,
    author = {Wei, Jason and Wang, Xuezhi and Schuurmans, Dale and Bosma, Maarten and Ichter, Brian and Xia, Fei and Chi, Ed H. and Le, Quoc V. and Zhou, Denny},
    title = {Chain-of-thought prompting elicits reasoning in large language models},
    year = {2022},
    isbn = {9781713871088},
    publisher = {Curran Associates Inc.},
    address = {Red Hook, NY, USA},
    booktitle = {Proceedings of the 36th International Conference on Neural Information Processing Systems},
    articleno = {1800},
    numpages = {14},
    location = {New Orleans, LA, USA},
    series = {NIPS '22}
}

@inproceedings{
    zelikman2022star,
    title={{ST}aR: Bootstrapping Reasoning With Reasoning},
    author={Eric Zelikman and Yuhuai Wu and Jesse Mu and Noah Goodman},
    booktitle={Advances in Neural Information Processing Systems},
    editor={Alice H. Oh and Alekh Agarwal and Danielle Belgrave and Kyunghyun Cho},
    year={2022},
    url={https://openreview.net/forum?id=_3ELRdg2sgI}
}

@misc{mukherjee2023orcaprogressivelearningcomplex,
      title={Orca: Progressive Learning from Complex Explanation Traces of GPT-4}, 
      author={Subhabrata Mukherjee and Arindam Mitra and Ganesh Jawahar and Sahaj Agarwal and Hamid Palangi and Ahmed Awadallah},
      year={2023},
      eprint={2306.02707},
      archivePrefix={arXiv},
      primaryClass={cs.CL},
      url={https://arxiv.org/abs/2306.02707}, 
}

@article{Guo_2025,
   title={DeepSeek-R1 incentivizes reasoning in LLMs through reinforcement learning},
   volume={645},
   ISSN={1476-4687},
   url={http://dx.doi.org/10.1038/s41586-025-09422-z},
   DOI={10.1038/s41586-025-09422-z},
   number={8081},
   journal={Nature},
   publisher={Springer Science and Business Media LLC},
   author={Guo, Daya and Yang, Dejian and Zhang, Haowei and Song, Junxiao and Wang, Peiyi and Zhu, Qihao and Xu, Runxin and Zhang, Ruoyu and Ma, Shirong and Bi, Xiao and Zhang, Xiaokang and Yu, Xingkai and Wu, Yu and Wu, Z. F. and Gou, Zhibin and Shao, Zhihong and Li, Zhuoshu and Gao, Ziyi and Liu, Aixin and Xue, Bing and Wang, Bingxuan and Wu, Bochao and Feng, Bei and Lu, Chengda and Zhao, Chenggang and Deng, Chengqi and Ruan, Chong and Dai, Damai and Chen, Deli and Ji, Dongjie and Li, Erhang and Lin, Fangyun and Dai, Fucong and Luo, Fuli and Hao, Guangbo and Chen, Guanting and Li, Guowei and Zhang, H. and Xu, Hanwei and Ding, Honghui and Gao, Huazuo and Qu, Hui and Li, Hui and Guo, Jianzhong and Li, Jiashi and Chen, Jingchang and Yuan, Jingyang and Tu, Jinhao and Qiu, Junjie and Li, Junlong and Cai, J. L. and Ni, Jiaqi and Liang, Jian and Chen, Jin and Dong, Kai and Hu, Kai and You, Kaichao and Gao, Kaige and Guan, Kang and Huang, Kexin and Yu, Kuai and Wang, Lean and Zhang, Lecong and Zhao, Liang and Wang, Litong and Zhang, Liyue and Xu, Lei and Xia, Leyi and Zhang, Mingchuan and Zhang, Minghua and Tang, Minghui and Zhou, Mingxu and Li, Meng and Wang, Miaojun and Li, Mingming and Tian, Ning and Huang, Panpan and Zhang, Peng and Wang, Qiancheng and Chen, Qinyu and Du, Qiushi and Ge, Ruiqi and Zhang, Ruisong and Pan, Ruizhe and Wang, Runji and Chen, R. J. and Jin, R. L. and Chen, Ruyi and Lu, Shanghao and Zhou, Shangyan and Chen, Shanhuang and Ye, Shengfeng and Wang, Shiyu and Yu, Shuiping and Zhou, Shunfeng and Pan, Shuting and Li, S. S. and Zhou, Shuang and Wu, Shaoqing and Yun, Tao and Pei, Tian and Sun, Tianyu and Wang, T. and Zeng, Wangding and Liu, Wen and Liang, Wenfeng and Gao, Wenjun and Yu, Wenqin and Zhang, Wentao and Xiao, W. L. and An, Wei and Liu, Xiaodong and Wang, Xiaohan and Chen, Xiaokang and Nie, Xiaotao and Cheng, Xin and Liu, Xin and Xie, Xin and Liu, Xingchao and Yang, Xinyu and Li, Xinyuan and Su, Xuecheng and Lin, Xuheng and Li, X. Q. and Jin, Xiangyue and Shen, Xiaojin and Chen, Xiaosha and Sun, Xiaowen and Wang, Xiaoxiang and Song, Xinnan and Zhou, Xinyi and Wang, Xianzu and Shan, Xinxia and Li, Y. K. and Wang, Y. Q. and Wei, Y. X. and Zhang, Yang and Xu, Yanhong and Li, Yao and Zhao, Yao and Sun, Yaofeng and Wang, Yaohui and Yu, Yi and Zhang, Yichao and Shi, Yifan and Xiong, Yiliang and He, Ying and Piao, Yishi and Wang, Yisong and Tan, Yixuan and Ma, Yiyang and Liu, Yiyuan and Guo, Yongqiang and Ou, Yuan and Wang, Yuduan and Gong, Yue and Zou, Yuheng and He, Yujia and Xiong, Yunfan and Luo, Yuxiang and You, Yuxiang and Liu, Yuxuan and Zhou, Yuyang and Zhu, Y. X. and Huang, Yanping and Li, Yaohui and Zheng, Yi and Zhu, Yuchen and Ma, Yunxian and Tang, Ying and Zha, Yukun and Yan, Yuting and Ren, Z. Z. and Ren, Zehui and Sha, Zhangli and Fu, Zhe and Xu, Zhean and Xie, Zhenda and Zhang, Zhengyan and Hao, Zhewen and Ma, Zhicheng and Yan, Zhigang and Wu, Zhiyu and Gu, Zihui and Zhu, Zijia and Liu, Zijun and Li, Zilin and Xie, Ziwei and Song, Ziyang and Pan, Zizheng and Huang, Zhen and Xu, Zhipeng and Zhang, Zhongyu and Zhang, Zhen},
   year={2025},
   month=Sept, pages={633–638}
}

@ARTICLE{11151751,
  author={Luo, Yun and Yang, Zhen and Meng, Fandong and Li, Yafu and Zhou, Jie and Zhang, Yue},
  journal={IEEE Transactions on Audio, Speech and Language Processing}, 
  title={An Empirical Study of Catastrophic Forgetting in Large Language Models During Continual Fine-Tuning}, 
  year={2025},
  volume={33},
  number={},
  pages={3776-3786},
  doi={10.1109/TASLPRO.2025.3606231}
}

@article{
    biderman2024lora,
    title={Lo{RA} Learns Less and Forgets Less},
    author={Dan Biderman and Jacob Portes and Jose Javier Gonzalez Ortiz and Mansheej Paul and Philip Greengard and Connor Jennings and Daniel King and Sam Havens and Vitaliy Chiley and Jonathan Frankle and Cody Blakeney and John Patrick Cunningham},
    journal={Transactions on Machine Learning Research},
    issn={2835-8856},
    year={2024},
    url={https://openreview.net/forum?id=aloEru2qCG},
    note={Featured Certification}
}

@inproceedings{mallen-etal-2023-trust,
    title = "When Not to Trust Language Models: Investigating Effectiveness of Parametric and Non-Parametric Memories",
    author = "Mallen, Alex  and
      Asai, Akari  and
      Zhong, Victor  and
      Das, Rajarshi  and
      Khashabi, Daniel  and
      Hajishirzi, Hannaneh",
    editor = "Rogers, Anna  and
      Boyd-Graber, Jordan  and
      Okazaki, Naoaki",
    booktitle = "Proceedings of the 61st Annual Meeting of the Association for Computational Linguistics (Volume 1: Long Papers)",
    month = jul,
    year = "2023",
    address = "Toronto, Canada",
    publisher = "Association for Computational Linguistics",
    url = "https://aclanthology.org/2023.acl-long.546/",
    doi = "10.18653/v1/2023.acl-long.546",
    pages = "9802--9822"
}

@inproceedings{
    10.1145/3600006.3613165,
    author = {Kwon, Woosuk and Li, Zhuohan and Zhuang, Siyuan and Sheng, Ying and Zheng, Lianmin and Yu, Cody Hao and Gonzalez, Joseph and Zhang, Hao and Stoica, Ion},
    title = {Efficient Memory Management for Large Language Model Serving with PagedAttention},
    year = {2023},
    isbn = {9798400702297},
    publisher = {Association for Computing Machinery},
    address = {New York, NY, USA},
    url = {https://doi.org/10.1145/3600006.3613165},
    doi = {10.1145/3600006.3613165},
    booktitle = {Proceedings of the 29th Symposium on Operating Systems Principles},
    pages = {611–626},
    numpages = {16},
    location = {Koblenz, Germany},
    series = {SOSP '23}
}

@misc{toolsense2026,
      title={ToolSense: A Diagnostic Framework for Auditing Parametric Tool Knowledge in LLMs}, 
      author={Ashutosh Hathidara and Sai Shruthi Sistla and Sebastian Schreiber and Sahil Bansal},
      year={2026},
      eprint={2606.12451},
      archivePrefix={arXiv},
      primaryClass={cs.AI},
      url={https://arxiv.org/abs/2606.12451}, 
}

@techreport{anthropic2026claudesonnet46,
  author      = {{Anthropic}},
  title       = {Claude Sonnet 4.6 System Card},
  institution = {Anthropic},
  year        = {2026},
  month       = {February},
  url         = {https://www-cdn.anthropic.com/bbd8ef16d70b7a1665f14f306ee88b53f686aa75/Claude%20Sonnet%204.6%20System%20Card.pdf}
}

@inproceedings{karpukhin2020dpr,
    title = "Dense Passage Retrieval for Open-Domain Question Answering",
    author = "Karpukhin, Vladimir  and
      Oguz, Barlas  and
      Min, Sewon  and
      Lewis, Patrick  and
      Wu, Ledell  and
      Edunov, Sergey  and
      Chen, Danqi  and
      Yih, Wen-tau",
    editor = "Webber, Bonnie  and
      Cohn, Trevor  and
      He, Yulan  and
      Liu, Yang",
    booktitle = "Proceedings of the 2020 Conference on Empirical Methods in Natural Language Processing (EMNLP)",
    month = nov,
    year = "2020",
    address = "Online",
    publisher = "Association for Computational Linguistics",
    url = "https://aclanthology.org/2020.emnlp-main.550/",
    doi = "10.18653/v1/2020.emnlp-main.550",
    pages = "6769--6781"
}

@misc{googledeepmind2026gemma4e4bit,
  title        = {{Gemma 4 E4B Instruction-Tuned Model}},
  author       = {{Google DeepMind}},
  year         = {2026},
  month        = {April},
  howpublished = {\url{https://huggingface.co/google/gemma-4-E4B-it}},
  note         = {Model overview: \url{https://ai.google.dev/gemma/docs/core/model_card_4}.
                  License: Apache 2.0},
}

\appendix

\section{Multi-Format Tool Memorization}
\label{app:multi-format-memo}

For the flat-token format we train on three formats jointly: (i)~the forward mapping $\mathrm{desc}{\to}v_t$, which grounds the description in the tool token; (ii)~the reverse mapping $v_t{\to}\mathrm{desc}$, which forces the model to recover the description from the token; and (iii)~a discriminative Multi-Choice Tool Selection (MCTS) objective in which, given $\mathrm{desc}_t$, the model must select the gold token $v_t$ from a contrast set of $K{+}1$ candidates ($v_t$ plus $K$ hard negatives mined from semantically adjacent tools). The reverse and MCTS formats together break the shortcut of name-to-token surface memorization and force genuine description-token alignment. For the hierarchical-token formats, we add two formats that target the API/endpoint factorization: $\mathrm{desc}{\to}v^{\mathrm{api}}_t$ and $(\mathrm{desc}, v^{\mathrm{api}}_t){\to}v^{\mathrm{endpoint}}_t$.

\section{Stage 2 Data Synthesis: Full Pipeline}
\label{app:stage2-pipeline}

This appendix provides the mechanics deferred from \S\ref{subsec:stage2-data}.

\paragraph{Hard-negative pools.}
Let $\phi : \mathcal{C} \to \mathbb{R}^d$ be a sentence encoder. For a tool $t$ we retrieve the top-$K$ neighbours under cosine similarity,
\begin{equation}
\mathcal{H}_K(t) = \operatorname*{arg\,top\text{-}K}_{t' \in \mathcal{C} \setminus \{t\}} \big\langle \phi(t), \phi(t') \big\rangle.
\end{equation}
The RRB anchor pool is $\mathcal{P}(t) = \{t\} \cup \mathcal{H}_K(t)$. The rule pool is
\begin{equation}
\mathcal{P}(r) \;=\; \mathcal{T}(r) \,\cup \bigcup_{t \in \mathcal{T}(r)} \mathcal{H}_K(t),
\end{equation}
which, by construction, contains every tool whose surface form is close enough to a tool in $\mathcal{T}(r)$ to make the problem more challenging --- thereby forcing the model to learn finer details of these tools.

\paragraph{Tier definitions (RRB).}
Three tier-wise sub-pipelines run in parallel for each anchor $t$, sampled stratified by domain. Each prompts the generator $\theta_{\mathrm{RRB}}$ with $(t, \mathcal{P}(t), \mathrm{tier}, \mathcal{E})$, where $\mathcal{E}$ is a small set of dynamic few-shot style real user queries to ground the generation towards realistic query distribution: \emph{easy} ($|A|{=}1$, fully-specified intent), \emph{medium} ($|A|\!\in\!\{2,3\}$, mild ambiguity), and \emph{hard} ($|A|\!\geq\!4$, multi-intent or under-specified). This tiered structure exposes the model to the ambiguity spectrum of real user queries rather than the verbose, fully-specified queries that we rarely encounter in production.

\paragraph{Rule-targeted generator.}
A separate generator $\theta_{\mathrm{rule}}$ receives $(r, \mathcal{P}(r), \mathcal{E})$ and is instructed to synthesize queries in three phrasing styles, each targeting a distinct mode in which a rule must be brought to bear at retrieval time:
\begin{itemize}
    \item \textbf{Explicit:} $q$ cites the rule verbatim or near-verbatim, supervising the model to recognize a rule when it is stated outright by the user.
    \item \textbf{Implicit:} $q$ alludes to the rule indirectly --- e.g., via a domain artifact (product line, year, region) that the rule conditions on --- without naming the rule itself, supervising the model to surface the rule from contextual cues.
    \item \textbf{Violation:} $q$ is deliberately under-specified so that, absent $r$, the answer set spans multiple tools in $\mathcal{T}(r)$; once $r$ is applied the answer collapses to a single tool. This is the style embedding retrievers cannot resolve at any threshold and the principal target of \trace's reasoning trace.
\end{itemize}
Each rule contributes a equal mix of the three styles, and every sample retains a pointer to its source rule $r^{\ast}$.

\paragraph{Trace generator.}
The teacher $\theta_{\mathrm{CoT}}$ receives the tuple $(q, A, \mathcal{P}, r^{\ast})$ --- where $r^{\ast} = \varnothing$ for RRB samples --- and follows the deliberative schema in \S\ref{subsec:stage2-data}: (i) restate the user intent; (ii) analyze the plausible candidates from $\mathcal{P}$ and articulate what distinguishes them; (iii) where applicable, state $r^{\ast}$ and apply it to eliminate confusables; (iv) commit to the answer set.

\paragraph{Programmatic filter $\mathcal{V}$.}
A sample is rejected if any of the following fail:
\begin{itemize}
\item \emph{grounding}: $A \subseteq \mathcal{P}$ and every tool referenced in $z$ exists in $\mathcal{P}$ (no hallucinated tools);
\item \emph{leakage}: no answer tool name appears verbatim in $q$, i.e., $\forall\, t' \in A : \mathrm{name}(t') \notin q$;
\item \emph{consistency}: the answer set committed at the end of $z$ exactly matches $y$, and $y$ is well-formed JSON over valid virtual tokens;
\item \emph{rule attribution}: for rule-targeted samples, $z$ explicitly invokes $r^{\ast}$.
\end{itemize}

\paragraph{LLM judge $\mathcal{J}$.}
Samples passing $\mathcal{V}$ are scored at temperature~0 on four axes: query naturalness, tier compliance, label correctness, and \emph{trace faithfulness} --- whether the reasoning in $z$ genuinely entails $A$ rather than rationalizing it post hoc. Rejected samples are regenerated with the judge's critique injected as feedback (up to a fixed retry budget of 5) and dropped thereafter.

\section{Training Setup}
\label{app:training}

Both Stage~1 and Stage~2 use standard autoregressive supervised fine-tuning. Each example is rendered as a chat-style prompt-completion pair, and the cross-entropy loss is masked over the user/system prompt tokens so that the model is supervised only on the assistant completion --- in Stage~2 this is the trace--answer pair $o = (\tilde{z},\, y)$ given the query $q$, and in Stage~1 the format-specific completion (e.g.\ the gold token $v_t$ in the $\mathrm{desc}{\to}v_t$ format). Optimizer, learning-rate schedule, batch size, sequence length, and LoRA rank/$\alpha$/dropout follow the ToolSense recipe~\citep{toolsense2026}.

\paragraph{Models and hyperparameters.}
All primary experiments use Gemma4-E4B-it with LoRA with embedding layers fully finetuned ($r=64$, $\alpha=128$) both for Stage~1 and Stage~2. Both stages use AdamW with cosine schedule (Stage~1: lr~$=$5e-5, Stage~2: lr~$=$~1e-4) with minimum LR ratio of 0.1, batch size of 8 for Stage~1 and 16 for Stage~2, bf16 precision, on a single H200 GPU.

\paragraph{Statistical reporting.}
All retrieval and probing metrics are reported with $95\%$ confidence intervals via paired bootstrap ($B{=}1{,}000$) over per-example predictions. MCQ$_\text{expert}$ is reported as a point estimate ($n{=}454$).

\section{Inference}
\label{app:inference}

At inference time \trace{} decodes greedily in a single forward pass: given a user query $q$, the model emits the trace $\hat{z}$ followed by the JSON token list $\hat{y} = [\, v_{a_1}, v_{a_2}, \ldots\,]$, terminating on the closing JSON bracket. We do not employ constrained beam search or prefix-trie expansion over the virtual-token vocabulary as in ToolGen~\citep{wang2025toolgen}; the parametric grounding from Stage~1 plus the deliberative scaffold from Stage~2 are sufficient to keep the answer list inside the valid token alphabet $\{v_t : t \in \mathcal{C}\}$ in practice \revision{(off-vocabulary rate of $1.04\%$/$1.96\%$ for Domain~A/B on \trace{} $(c, m, r{+}R)$; see Appendix~\ref{app:extended-metrics})}. The retrieved tool set is recovered as $\hat{A}(q) = \{t : v_t \in y\}$. \revision{In deployment, reasoning traces serve as internal scaffolds never surfaced to end users; only the retrieved tool set $\hat{A}(q)$ is passed to the downstream orchestrator. Off-vocabulary tokens (tokens not mapping to any valid tool in $\mathcal{C}$) are silently discarded.} This single-beam, no-constraint decoding is the source of \trace{}'s production-latency benefit: end-to-end retrieval cost is one prompt-prefill plus a short autoregressive generation, with no per-step trie lookup. \revision{Figure~\ref{fig:latency} quantifies this benefit for the headline results summarized in \S\ref{subsec:latency}.}

\begin{figure}[t]
  \centering
  \begin{subfigure}{0.48\linewidth}
    \includegraphics[width=\linewidth]{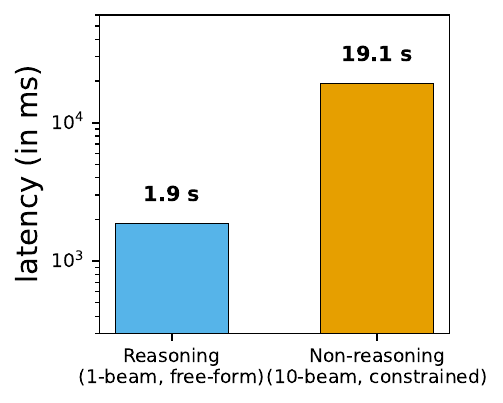}
    \caption{Single-user latency at batch=1.}
    \label{fig:latency-single}
  \end{subfigure}
  \hfill
  \begin{subfigure}{0.48\columnwidth}
    \includegraphics[width=\linewidth]{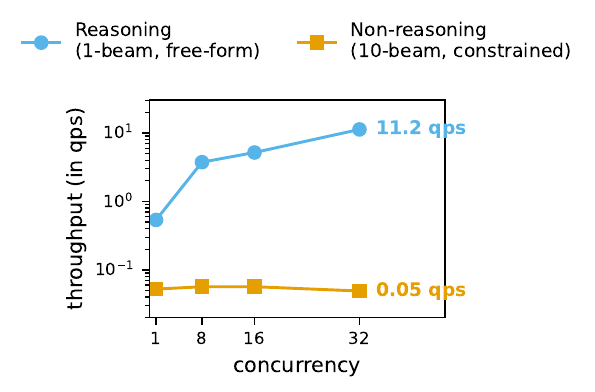}
    \caption{Throughput vs concurrency.}
    \label{fig:latency-throughput}
  \end{subfigure}
  \caption{\textbf{Latency comparison} of \trace{} $(c, m, r{+}R(12{-}rep))$ on two inference models using vLLM on a single H200, $100$ Domain-A PRB queries (log axes).}
  \label{fig:latency}
\end{figure}

\section{Detailed Evaluation Results}
\label{app:detailed-evaluation}

This section provides the full evaluation results for the experiments summarized in the main body, including confidence intervals and all metrics.

\subsection{Probing Evaluations}
\label{app_subsec:probing_evaluations}

This appendix tabulates per-format probing results for every \trace{} variant we trained, completing the results in \S\ref{subsec:preservation}. Throughout, runs are referenced by their configuration triple $(F, M, R)$ where $F$ is the token format (\textsf{a}~flat, \textsf{b}~bare-hier, \textsf{c}~wrapped-hier), $M$ is the Stage-1 mode (\textsf{m}~memo, \textsf{mr}~reasoning memo, $\varnothing$~none), and $R$ is the Stage-2 mode ($\varnothing$~none, \textsf{n}~non-reasoning retrieval, \textsf{n}$\to$\textsf{r}~non-reasoning then reasoning, \textsf{r}~reasoning, \textsf{r{+}R}~reasoning with rule grounding). All numbers are accuracy in \%; brackets are $95\%$ paired-bootstrap CIs ($B{=}1{,}000$). MCQ$_\text{expert}$ is reported as a point estimate.

\begin{table*}[t]
  \centering
  \small
  \resizebox{\textwidth}{!}{%
  \begin{tabular}{ccc ccccc}
    \toprule
    $F$ & $M$ & $R$ & MCQ$_\text{ts}$-A & MCQ$_\text{ts}$-B & QA$_\text{ts}$-A & QA$_\text{ts}$-B & MCQ$_\text{expert}$ \\
    \midrule
    \multirow{4}{*}{\textsf{a}} & \textsf{m} & $\varnothing$           & 52.3 [47.0,\,57.7] & 50.2 [45.5,\,55.5] & 80.0 [74.5,\,85.5] & 67.8 [62.7,\,72.2] & 63.4 \\
                                & \textsf{m} & \textsf{n}              & 27.3 [22.7,\,32.7] & 29.8 [25.2,\,34.2] & 51.5 [44.5,\,58.5] & 52.5 [47.5,\,57.0] & 34.8 \\
                                & \textsf{m} & \textsf{n}$\to$\textsf{r} & 65.0 [60.0,\,70.7] & 71.5 [67.2,\,76.0] & 76.5 [71.0,\,82.0] & 80.0 [76.0,\,83.8] & 49.8 \\
                                & \textsf{m} & \textsf{r}              & 68.7 [63.7,\,73.7] & 70.0 [65.5,\,74.5] & 87.5 [83.0,\,91.5] & 85.5 [82.0,\,88.8] & 59.5 \\
    \midrule
    \multirow{4}{*}{\textsf{b}} & \textsf{m} & $\varnothing$           & 43.3 [37.3,\,49.0] & 37.2 [32.8,\,42.0] & 61.5 [54.5,\,67.5] & 55.8 [50.7,\,60.8] & 58.6 \\
                                & \textsf{m} & \textsf{n}              & 28.3 [23.3,\,33.3] & 27.0 [22.5,\,31.8] & 45.5 [38.5,\,52.5] & 43.0 [38.2,\,48.0] & 30.0 \\
                                & \textsf{m} & \textsf{n}$\to$\textsf{r} & 28.7 [23.7,\,34.0] & 32.0 [27.5,\,36.5] & 80.5 [75.0,\,85.5] & 69.0 [64.5,\,73.2] & 28.2 \\
                                & \textsf{m} & \textsf{r}              & 46.3 [41.0,\,52.0] & 44.0 [39.2,\,48.8] & 84.0 [79.0,\,88.5] & 79.8 [75.8,\,83.5] & 50.9 \\
    \midrule
    \multirow{4}{*}{\textsf{c}} & \textsf{m} & $\varnothing$           & 60.0 [54.3,\,65.3] & 59.0 [54.0,\,63.7] & 75.0 [69.0,\,81.0] & 69.2 [64.8,\,74.0] & 69.2 \\
                                & \textsf{m} & \textsf{n}              & 24.3 [19.3,\,29.3] & 23.5 [19.5,\,27.5] & 50.0 [43.0,\,56.5] & 50.0 [45.2,\,54.8] & 33.7 \\
                                & \textsf{m} & \textsf{n}$\to$\textsf{r} & 24.7 [19.7,\,29.7] & 24.8 [20.2,\,28.7] & 50.0 [43.0,\,57.0] & 50.0 [45.5,\,54.8] & 31.5 \\
                                & \textsf{m} & \textsf{r}              & 57.7 [52.3,\,63.0] & 67.8 [63.2,\,72.2] & 83.0 [77.5,\,88.5] & 79.2 [75.5,\,83.0] & 61.5 \\
    \bottomrule
  \end{tabular}%
  }
  \caption{\textbf{Token-format sweep.} Probing accuracy across the three token
  formats and four Stage-2 modes, all with non-reasoning Stage-1 ($M{=}\textsf{m}$).
  Non-reasoning retrieval ($R{=}\textsf{n}$) collapses every probe across all
  formats; reasoning retrieval ($R{=}\textsf{r}$) recovers it. Bolting reasoning
  onto a non-reasoning checkpoint ($R{=}\textsf{n}{\to}\textsf{r}$) only partially
  recovers MCQ accuracy and fails entirely for $F{\in}\{\textsf{b},\textsf{c}\}$.}
  \label{tab:app-format-sweep}
\end{table*}

\textbf{Format sweep (Table~\ref{tab:app-format-sweep}).} The dissociation under $R{=}\textsf{n}$ is consistent across all three token formats: every $(\cdot, \textsf{m}, \textsf{n})$ row sits around the random baseline on all probes. Replacing $R{=}\textsf{n}$ with $R{=}\textsf{r}$ recovers the in-distribution probes to within a few points of the $(\cdot, \textsf{m}, \varnothing)$ ceiling for every $F$. The \textsf{n}$\to$\textsf{r} curriculum is informative: on flat tokens ($F{=}\textsf{a}$) the late reasoning bolt-on partially recovers MCQ; on hierarchical formats ($F{\in}\{\textsf{b},\textsf{c}\}$) it does not, suggesting that the damage caused by non-reasoning training to the description--token binding is irrecoverable when the token vocabulary itself carries structural content. Wrapped-hierarchical ($F{=}\textsf{c}$) achieves the highest $(\cdot, \textsf{m}, \varnothing)$ ceiling on MCQ$_\text{expert}$ ($69.2$), which is why we showcased the results corresponding to that format in the main body.

\begin{table}[t]
  \centering
  \small
  \resizebox{\columnwidth}{!}{%
  \begin{tabular}{ccc cc c}
    \toprule
    $F$ & $M$ & $R$ & MCQ$_\text{ts}$-A & MCQ$_\text{ts}$-B & MCQ$_\text{expert}$ \\
    \midrule
    \textsf{a} & $\varnothing$ & \textsf{r} & 74.0 [69.0,\,78.7] & 81.8 [78.0,\,85.2] & 67.8 \\
    \textsf{a} & \textsf{m}    & \textsf{r} & 68.7 [63.7,\,73.7] & 70.0 [65.5,\,74.5] & 59.5 \\
    \textsf{a} & \textsf{mr}   & \textsf{r} & 62.3 [57.0,\,68.0] & 73.5 [69.0,\,77.5] & 65.2 \\
    \midrule
    \textsf{b} & \textsf{m}    & \textsf{r} & 46.3 [41.0,\,52.0] & 44.0 [39.2,\,48.8] & 50.9 \\
    \textsf{b} & \textsf{mr}   & \textsf{r} & 62.0 [56.3,\,67.3] & 71.5 [67.0,\,76.0] & 63.2 \\
    \midrule
    \textsf{c} & \textsf{m}    & \textsf{r} & 57.7 [52.3,\,63.0] & 67.8 [63.2,\,72.2] & 61.5 \\
    \textsf{c} & \textsf{mr}   & \textsf{r} & 58.3 [53.0,\,63.7] & 67.8 [63.2,\,72.2] & 65.2 \\
    \bottomrule
  \end{tabular}%
  }
  \caption{\textbf{Stage-1 ablation under reasoning retrieval.} Holding
  $R{=}\textsf{r}$ fixed, we vary $M\in\{\varnothing, \textsf{m}, \textsf{mr}\}$
  to isolate Stage-1's contribution. The $(\textsf{a}, \varnothing, \textsf{r})$ row
  is the no-Stage-1 ablation: it recovers MCQ$_\text{expert}$ to within $1$\,pp
  of the base model, but has the lowest retrieval.
  Reasoning Stage-1 ($M{=}\textsf{mr}$) outperforms non-reasoning Stage-1
  ($M{=}\textsf{m}$) on MCQ$_\text{expert}$ for every $F$.}
  \label{tab:app-stage1-ablation}
\end{table}

\textbf{Stage-1 ablation (Table~\ref{tab:app-stage1-ablation}).} The $(\textsf{a}, \varnothing, \textsf{r})$ row is striking: training reasoning retrieval directly from the base model (no Stage-1) recovers MCQ$_\text{expert}$ to $67.8$ --- within $1.4$\,pp of the strongest Stage-1 ceiling $(\textsf{c}, \textsf{m}, \varnothing) = 69.2$ --- but its retrieval recall is the lowest of any \trace{} variant. This suggests that Stage-1 alone is not responsible for building tool understanding. Switching $M{=}\textsf{m}\to\textsf{mr}$ (reasoning at Stage-1) yields a consistent $2$--$13$\,pp gain on MCQ$_\text{expert}$ across formats, suggesting that reasoning at the memorization stage helps in improving tool understanding.

\begin{table}[t]
  \centering
  \tiny
  \begin{tabular}{ccc ccc}
    \toprule
    $F$ & $M$ & $R$ & MCQ$_\text{ts}$-A & MCQ$_\text{ts}$-B & MCQ$_\text{expert}$ \\
    \midrule
    \textsf{a} & \textsf{m} & \textsf{r}                  & 68.7 [63.7,\,73.7] & 70.0 [65.5,\,74.5] & 59.5 \\
    \textsf{a} & \textsf{m} & \textsf{r}{+}\textsf{R}(3q) & 66.7 [61.3,\,72.0] & 62.5 [57.7,\,67.2] & 59.2 \\
    \textsf{a} & \textsf{m} & \textsf{r}{+}\textsf{R}(6q) & 68.0 [63.0,\,73.7] & 66.2 [61.7,\,70.8] & 58.1 \\
    \textsf{a} & \textsf{m} & \textsf{r}{+}\textsf{R}(9q) & 66.3 [61.0,\,71.7] & 62.0 [57.5,\,67.0] & 57.9 \\
    \textsf{a} & \textsf{m} & \textsf{r}{+}\textsf{R}(\textit{app}) & 62.3 [56.3,\,67.7] & 66.8 [61.8,\,71.5] & --- \\
    \textsf{a} & \textsf{m} & \textsf{r}{+}\textsf{R}(\textit{rep}) & 65.7 [60.3,\,71.0] & 63.0 [58.2,\,68.0] & 56.2 \\
    \textsf{a} & \textsf{m} & \textsf{r}{+}\textsf{R}             & 58.7 [53.0,\,64.3] & 68.0 [63.2,\,72.3] & 55.7 \\
    \midrule
    \textsf{c} & \textsf{m} & \textsf{r}                  & 57.7 [52.3,\,63.0] & 67.8 [63.2,\,72.2] & 61.5 \\
    \textsf{c} & \textsf{m} & \textsf{r}{+}\textsf{R}     & 56.7 [51.0,\,62.3] & 71.0 [66.8,\,75.2] & 56.4 \\
    \bottomrule
  \end{tabular}
  \caption{\textbf{Rule-grounded scaling.} Probing accuracy along the rule-data
  axis. Rows above the rule entries reproduce the no-rules baseline
  $(F, \textsf{m}, \textsf{r})$ for direct comparison. Held-out
  MCQ$_\text{expert}$ falls monotonically with rule-data volume; in-distribution
  probes are stable.}
  \label{tab:app-rule-scaling}
\end{table}

\textbf{Rule-grounded scaling (Table~\ref{tab:app-rule-scaling}).} The rule-data axis is reported in two layers: the upper block sweeps $(F{=}\textsf{a}, \textsf{m}, \textsf{r}{+}\textsf{R}(\cdot))$ across rule-data volumes ($k$\,q/rule, append vs.\ replace strategies); the lower block reproduces the headline $(F{=}\textsf{c}, \textsf{m}, \cdot)$ pair for direct comparison with the body. Two trends are visible. First, in-distribution probes are stable across the rule-data axis: MCQ$_\text{ts}$ and QA$_\text{ts}$ accuracy moves within their own CIs as we scale rule-data volume. Second, MCQ$_\text{expert}$ exhibits the Pareto cost: each rule-data increment costs roughly $1$--$2$\,pp of held-out general-domain knowledge, with the lowest landing at $55.7$ ($F{=}\textsf{a}$) and $56.4$ ($F{=}\textsf{c}$), still well above the $R{=}\textsf{n}$ collapse but below the no-rules ceiling. The $(\textit{app})$ row's missing MCQ$_\text{expert}$ entry is because the corresponding checkpoint has not yet been evaluated.

\subsection{Retrieval Evaluations}
\label{app_subsec:retrieval_evals}

Tables~\ref{tab:app-headhead-vb} and~\ref{tab:app-headhead-va} tabulate the full set of retrieval results for every $(F, M, R)$ checkpoint we trained, completing the comparisons in \S\ref{subsec:retrieval}. Each checkpoint is reported under whichever decoding mode it was evaluated with (non-reasoning checkpoints under constrained decoding, reasoning checkpoints under single-beam reasoning decoding); for the two \trace{}-class checkpoints that we additionally re-evaluated under constrained decoding (suffix \textit{cstr}, marked with $^\ast$), both decoding modes appear. Numbers are recall in \%; brackets are $95\%$ paired-bootstrap CIs ($B{=}1{,}000$) over per-example predictions.

\begin{table*}[t]
  \centering\small
  \resizebox{\textwidth}{!}{%
  \begin{tabular}{ccc cc cc cc}
    \toprule
    & & & \multicolumn{3}{c}{Domain~A} & \multicolumn{3}{c}{Domain~B} \\
    \cmidrule(lr){4-6}\cmidrule(lr){7-9}
    $F$ & $M$ & $R$ & R@1 & R@5 & R@10 & R@1 & R@5 & R@10 \\
    \midrule
    \multicolumn{9}{l}{\emph{Non-reasoning retrieval baselines}} \\
    \textsf{a} & \textsf{m} & \textsf{n}             & 39.7 \,[31.3,\,48.1] & 80.9 \,[74.0,\,87.8] & 91.6 \,[86.3,\,96.2] & 31.7 \,[23.6,\,40.7] & 69.9 \,[61.8,\,77.2] & 80.5 \,[73.2,\,87.0] \\
    \textsf{b} & \textsf{m} & \textsf{n}             & 44.3 \,[35.9,\,52.7] & 80.2 \,[73.3,\,86.3] & 86.3 \,[80.2,\,91.6] & 40.7 \,[31.7,\,49.6] & 70.7 \,[63.4,\,77.2] & 79.7 \,[73.2,\,86.2] \\
    \textsf{c} & \textsf{m} & \textsf{n}             & 45.0 \,[36.6,\,54.2] & 80.9 \,[74.0,\,87.0] & 87.0 \,[80.9,\,92.4] & 35.8 \,[27.6,\,43.9] & 73.2 \,[65.9,\,80.5] & 78.0 \,[70.7,\,85.4] \\
    \midrule
    \multicolumn{9}{l}{\emph{\trace{}-class re-evaluations under constrained decoding (suffix $^\ast$)}} \\
    \textsf{c} & \textsf{m} & \textsf{r}$^\ast$         & 40.5 \,[32.1,\,48.9] & 67.2 \,[58.8,\,75.6] & 74.0 \,[66.4,\,81.7] & 46.3 \,[37.4,\,54.5] & 75.6 \,[67.5,\,82.9] & 79.7 \,[72.4,\,87.0] \\
    \textsf{a} & \textsf{m} & \textsf{r}{+}\textsf{R}$^\ast$ & 24.4 \,[16.8,\,32.1] & 44.3 \,[35.9,\,52.7] & 54.2 \,[45.8,\,62.6] & 25.2 \,[17.9,\,33.3] & 52.8 \,[43.9,\,61.8] & 65.0 \,[56.9,\,73.2] \\
    \textsf{c} & \textsf{m} & \textsf{r}{+}\textsf{R}$^\ast$ & 48.9 \,[40.5,\,57.3] & 78.6 \,[71.0,\,85.5] & 86.3 \,[79.4,\,91.6] & 36.6 \,[27.6,\,45.5] & 69.1 \,[60.2,\,77.2] & 77.2 \,[69.9,\,84.6] \\
    \bottomrule
  \end{tabular}}
  \caption{\textbf{Constrained-decoded retrieval (R@1, R@5, R@10).}
  Includes the non-reasoning retrieval baselines $(\cdot, \textsf{m},
  \textsf{n})$ across all three token formats and the apple-to-apple
  re-evaluations (suffix $^\ast$) of two \trace{}-class reasoning checkpoints.}
  \label{tab:app-headhead-vb}
\end{table*}

\begin{table*}[t]
  \centering\small
  \resizebox{\textwidth}{!}{%
  \begin{tabular}{ccc cc cc}
    \toprule
    & & & \multicolumn{2}{c}{Domain~A} & \multicolumn{2}{c}{Domain~B} \\
    \cmidrule(lr){4-5}\cmidrule(lr){6-7}
    $F$ & $M$ & $R$ & R@1 & R@gen & R@1 & R@gen \\
    \midrule
    \multicolumn{7}{l}{\emph{Non-reasoning then reasoning $(M{=}\textsf{m}, R{=}\textsf{n}{\to}\textsf{r})$}} \\
    \textsf{a} & \textsf{m} & \textsf{n}$\to$\textsf{r} & 44.3 \,[35.9,\,53.4] & 58.0 \,[49.6,\,66.4] & 44.7 \,[35.8,\,53.7] & 52.0 \,[43.1,\,61.0] \\
    \textsf{b} & \textsf{m} & \textsf{n}$\to$\textsf{r} & 48.9 \,[40.5,\,57.3] & 61.8 \,[53.4,\,69.5] & 56.9 \,[48.0,\,65.9] & 63.4 \,[54.5,\,71.5] \\
    \textsf{c} & \textsf{m} & \textsf{n}$\to$\textsf{r} & 34.4 \,[26.7,\,42.7] & 45.0 \,[36.6,\,53.4] & 48.0 \,[39.0,\,56.9] & 54.5 \,[45.5,\,63.4] \\
    \midrule
    \multicolumn{7}{l}{\emph{Reasoning retrieval, no rules $(M{=}\textsf{m}, R{=}\textsf{r})$}} \\
    \textsf{a} & \textsf{m} & \textsf{r}              & 38.2 \,[29.8,\,46.6] & 55.7 \,[46.6,\,65.6] & 55.3 \,[46.3,\,64.2] & 62.6 \,[53.7,\,70.7] \\
    \textsf{b} & \textsf{m} & \textsf{r}              & 30.5 \,[22.9,\,38.2] & 46.6 \,[38.2,\,55.0] & 54.5 \,[45.5,\,63.4] & 60.2 \,[51.2,\,68.3] \\
    \textsf{c} & \textsf{m} & \textsf{r}              & 32.1 \,[24.4,\,40.5] & 53.4 \,[45.0,\,61.1] & 56.1 \,[47.2,\,65.0] & 67.5 \,[59.3,\,74.8] \\
    \midrule
    \multicolumn{7}{l}{\emph{Reasoning memo + reasoning retrieval $(M{=}\textsf{mr}, R{=}\textsf{r})$}} \\
    \textsf{a} & \textsf{mr} & \textsf{r}             & 31.3 \,[23.7,\,39.0] & 41.2 \,[33.6,\,49.6] & 44.7 \,[35.8,\,53.7] & 61.0 \,[51.2,\,69.9] \\
    \textsf{b} & \textsf{mr} & \textsf{r}             & 42.0 \,[33.6,\,50.4] & 52.7 \,[44.3,\,61.1] & 52.0 \,[43.1,\,61.0] & 65.0 \,[56.1,\,73.2] \\
    \textsf{c} & \textsf{mr} & \textsf{r}             & 41.2 \,[33.6,\,49.6] & 49.6 \,[41.2,\,58.0] & 53.7 \,[44.7,\,62.6] & 62.6 \,[53.7,\,71.5] \\
    \midrule
    \multicolumn{7}{l}{\emph{Direct reasoning, no Stage-1 $(M{=}\varnothing, R{=}\textsf{r})$}} \\
    \textsf{a} & $\varnothing$ & \textsf{r}           & 29.8 \,[22.1,\,37.4] & 45.8 \,[37.4,\,54.2] & 46.3 \,[37.4,\,55.3] & 61.8 \,[52.8,\,70.7] \\
    \midrule
    \multicolumn{7}{l}{\emph{Rule-grounded reasoning retrieval on $F{=}\textsf{a}$ scaling sweep}} \\
    \textsf{a} & \textsf{m} & \textsf{r}{+}\textsf{R}(3q)            & 25.2 \,[17.6,\,32.8] & 48.9 \,[40.5,\,57.3] & 59.3 \,[50.4,\,68.3] & 65.0 \,[56.1,\,73.2] \\
    \textsf{a} & \textsf{m} & \textsf{r}{+}\textsf{R}(6q)            & 42.0 \,[33.6,\,50.4] & 57.3 \,[48.9,\,65.6] & 52.0 \,[43.1,\,61.0] & 61.8 \,[52.8,\,70.7] \\
    \textsf{a} & \textsf{m} & \textsf{r}{+}\textsf{R}(9q)            & 38.9 \,[30.5,\,47.3] & 57.3 \,[48.9,\,65.6] & 68.3 \,[59.3,\,77.2] & 72.4 \,[64.2,\,80.5] \\
    \textsf{a} & \textsf{m} & \textsf{r}{+}\textsf{R}(12)            & 45.8 \,[37.4,\,54.2] & 73.3 \,[65.6,\,80.9] & 62.6 \,[53.7,\,71.5] & 65.9 \,[56.9,\,74.0] \\
    \textsf{a} & \textsf{m} & \textsf{r}{+}\textsf{R}(12{-}rep) & 44.3 \,[35.9,\,52.7] & 78.6 \,[71.8,\,84.7] & 52.8 \,[43.9,\,61.8] & 61.8 \,[52.8,\,70.7] \\
    \textsf{a} & \textsf{m} & \textsf{r}{+}\textsf{R}                & 48.1 \,[40.5,\,57.3] & 86.3 \,[80.9,\,91.6] & 61.0 \,[52.0,\,69.9] & 64.2 \,[56.1,\,72.4] \\
    \midrule
    \multicolumn{7}{l}{\emph{\trace{} on $F{=}\textsf{c}$ (headline)}} \\
    \textsf{c} & \textsf{m} & \textsf{r}{+}\textsf{R}                & 51.1 \,[42.7,\,60.3] & 85.5 \,[79.4,\,90.8] & 43.1 \,[35.0,\,52.1] & 60.2 \,[51.2,\,69.1] \\
    \bottomrule
  \end{tabular}}
  \caption{\textbf{Single-beam reasoning retrieval (R@1, R@gen).} Each row
  is a checkpoint evaluated by letting the model commit to its own answer
  set; R@gen is recall over the generated set. Sections group runs by
  $(M, R)$.}
  \label{tab:app-headhead-va}
\end{table*}




\subsection{Qualitative Trace Example}
\label{app_subsec:qualitative}

Figure~\ref{fig:qualitative_trace_app} shows a representative side-by-side comparison of model reasoning traces before and after rule grounding, illustrating how the \texttt{<think>} trace mechanism enables explicit rule citation for ambiguous queries. The left panel shows the ungrounded model reasoning plausibly but arriving at the wrong tool; the right panel shows how citing the governing business rule within the trace corrects the retrieval.

\begin{figure*}[t]
\centering
\tcbset{raster equal height=rows}
\begin{tcbraster}[raster columns=2, raster equal height, raster column skip=0.5cm, raster width=\linewidth]
\begin{tcolorbox}[colback=red!4!white, colframe=red!50!black,
    title={\textbf{Before Rule Grounding}}, fonttitle=\small\bfseries]
  \textbf{Query:} \textit{``Show my cost center details''}\\[4pt]
  {\small\ttfamily <think>\\
  The user wants cost center details. \ldots\\
  Cost centers relate to org units. WorkAssignment has org unit fields.\\
  \ldots
  </think>}\\[4pt] \\
  \textbf{Output:} {\small\ttfamily ["<<User\&\&WorkAssignment>>"]}\\[4pt]
  {\color{red}\textbf{$\times$ Incorrect}} -- wrong entity selected
\end{tcolorbox}
\begin{tcolorbox}[colback=green!4!white, colframe=green!50!black,
    title={\textbf{After Rule Grounding}}, fonttitle=\small\bfseries]
  \textbf{Query:} \textit{``Show my cost center details''}\\[4pt]
  {\small\ttfamily <think>\\
  The user wants cost center details. \ldots\\
  Per MASTER\_DATA rule: employee-attribute\\
  queries incl.\ cost center $\rightarrow$ EmployeeInfo.\\
  WorkAssignment does not hold cost center.\\
  \ldots
  </think>}\\[4pt]
  \textbf{Output:} {\small\ttfamily ["<<Employee\&\&EmployeeInfo>>"]}\\[4pt]
  {\color{green!60!black}\textbf{$\checkmark$ Correct}} -- rule citation guides routing
\end{tcolorbox}
\end{tcbraster}
\caption{\textbf{Rule grounding corrects ambiguous tool routing.} Without rule grounding (left), the model reasons plausibly but selects the wrong tool because it lacks the domain routing constraint. After rule grounding (right), the reasoning trace explicitly cites the business rule, producing the correct retrieval.}
\label{fig:qualitative_trace_app}
\end{figure*}

\section{MCQ$_\text{expert}$ Evaluation Details}
\label{app_subsec:eval-set-construction-mcq-expert}
MCQ$_\text{expert}$ is a multiple-choice question dataset handcrafted by domain experts, 
used to evaluate model performance on Domain~A and Domain~B knowledge.
\subsection{Dataset Construction}
We acquire a large dataset of multiple-choice questions and answers written by 
domain experts, used to assess domain understanding when preparing for certification 
exams. Each question belongs to a lesson group which is a curated set of instructional 
materials such as videos and documentation. We retain only questions relevant to 
Domain~A and Domain~B by filtering on keywords `A' and `B' in the lesson group title, 
followed by de-duplication to remove questions appearing in multiple lessons. This 
leaves 454 multiple-choice questions, each with 2 to 5 answer choices and exactly one 
correct answer, with the distribution of answer choices shown in Table~\ref{tab:mcq_distribution}.
\begin{table}[ht]
\centering
\begin{tabular}{cc}
\toprule
\textbf{No. of Choices} & \textbf{No. of Questions} \\
\midrule
2 & 58 \\
3 & 74 \\
4 & 301 \\
5 & 21 \\
\midrule
Total & 454 \\
\bottomrule
\end{tabular}
\caption{Distribution of answer choices in the MCQ$_\text{expert}$ dataset.}
\label{tab:mcq_distribution}
\end{table}
The random chance baseline is computed as $\frac{1}{N}\displaystyle\sum_{i=1}^{N} 
\frac{1}{k_i}$, where $N = 454$ and $k_i$ is the number of choices for question $i$. 
With $n_k$ denoting the number of questions with $k$ choices (Table~\ref{tab:mcq_distribution}), this gives:
\begin{equation}
\begin{split}
    \text{MCQ}_\text{expert}^{\text{random}} &= \frac{1}{N} \displaystyle\sum_{k=2}^{5} \frac{n_k}{k} \\
    &= \frac{1}{454} \left( \frac{58}{2} + \frac{74}{3} + \frac{301}{4} + \frac{21}{5} \right) \\
    &\approx 0.2932
\end{split}
\end{equation}
\subsection{Model Inference}
During inference, we prompt the model with the question content and the answer choices, 
where each answer choice is prefixed by the corresponding indexed letter of the English 
alphabet. We perform greedy constrained decoding, restricting the output vocabulary to 
the set of valid answer letters. Formally, given a question with $k$ choices, the 
model's prediction is:
\begin{equation}
    \hat{a} = \underset{l \in \mathcal{A}_k}{\arg\max} \ P(l \mid \text{prompt})
\end{equation}
where $\mathcal{A}_k = \{\texttt{A}, \texttt{B}, \dots\}$ is the set of $k$ valid 
answer letters and $P(l \mid \text{prompt})$ is the model's predicted probability 
for letter $l$. The model is evaluated by comparing $\hat{a}$ to the ground truth 
label $a^*$, and accuracy is reported as the fraction of correct predictions over 
all $N$ questions.
\subsection{Example Question}
The following example is selected at random from the dataset of 454 questions. 
\begin{tcolorbox}[colback=gray!5!white, colframe=gray!60!black, fonttitle=\small\bfseries, title={MCQ$_\text{expert}$ Example Question}]
\begin{lstlisting}[basicstyle=\small\ttfamily, breaklines=true, breakatwhitespace=true, columns=fullflexible]
Answer the following question. 
Respond with the answer letter only.

Which of the following steps is NOT 
typically part of the basic financing 
process for indirect materials?
A: Requirements Determination
B: Source of Supply Determination
C: Production Planning and Control
D: Purchase Order Processing
Answer (single letter only): 
\end{lstlisting}
\end{tcolorbox}
The correct answer is \texttt{C}.


\section{Business Rule Catalog}
\label{app:business-rules}

The business rule catalog $\mathcal{B}$ comprises \textbf{123 rules} across the two domains: 20 rules for Domain~A  and 103 rules for Domain~B . Each rule is authored by domain experts and specifies how to disambiguate between a set of semantically overlapping tools (the \emph{confusible set} $\mathcal{T}(r)$). Rules capture API deprecations, versioning constraints, product-line routing logic, and domain-specific data-model boundaries that are not inferable from tool descriptions alone.

\subsection{Rule Structure}

Each rule $r \in \mathcal{B}$ is stored as a JSON object with three fields: \textbf{\texttt{tool\_name}} (the target tool the rule governs), \textbf{\texttt{confusables}} (the set of tools $\mathcal{T}(r)$ that overlap semantically with the target), and \textbf{\texttt{rule\_text}} (a natural-language directive specifying when to route to this tool vs.\ its confusables).

\subsection{Example Rule}

The following example from Domain~A illustrates a business rule.

\begin{tcolorbox}[colback=gray!5!white, colframe=gray!60!black, fonttitle=\small\bfseries, title={Example Rule}]
\begin{lstlisting}[basicstyle=\small\ttfamily, breaklines=true, breakatwhitespace=true, columns=fullflexible]
{
  "tool_name": "API1/EndpointA",
  "confusables": [
    "API1/EndpointB",
    "API1/EndpointC",
    "API1/EndpointD"
  ],
  "rule_text": "EndpointA is the default for current-state queries that do not require historical data. Route to EndpointB when the query involves historical records or date-range tracking. Route to EndpointC when the query targets transactional or line-item level details, and to EndpointD otherwise."
}
\end{lstlisting}
\end{tcolorbox}

\subsection{Rule Statistics}

\begin{table}[H]
\centering
\footnotesize
\begin{tabular}{lcc}
\toprule
& Domain~A & Domain~B \\
\midrule
Rules & 20 & 103 \\
Avg.\ confusables & 4.2 & 2.8 \\
Avg.\ length (words) & $\sim$180 & $\sim$120 \\
\bottomrule
\end{tabular}
\caption{Business rule catalog statistics.}
\label{tab:rule-stats}
\end{table}

\section{System Prompts}
\label{app:system-prompts}

This section reproduces the system prompts used in the Stage~2 data synthesis pipeline (\S\ref{subsec:stage2-data}). All prompts use Jinja2 templating; variables in double braces (e.g., \texttt{\{\{tool\_name\}\}}) are filled at runtime.

\subsection{Rule-Targeted Query Generation}
\label{app:prompt-query-gen}

The following prompt generates queries in three categories (explicit, implicit, exception) for a given business rule. It is invoked once per (rule, category) pair..

\begin{tcolorbox}[colback=gray!3!white, colframe=gray!50!black, fonttitle=\small\bfseries, title={Prompt: Rule-Targeted Query Generator}, breakable]
\small\ttfamily
You are tasked at generating diverse user queries to evaluate a business routing rule for a specific tool.\\[4pt]
<BUSINESS RULE>\\
Tool: \{\{tool\_name\}\}\\
Rule: \{\{rule\_name\}\}\\
\{\{rule\_text\}\}\\
</BUSINESS RULE>\\[4pt]
<TOOL DESCRIPTION>\\
\{\{tool\_description\}\}\\
</TOOL DESCRIPTION>\\[4pt]
<CATEGORY>\\
Generate queries in the "\{\{category\}\}" category:\\[2pt]
- \textbf{explicit}: The query wording makes it obvious that the business rule applies. Keywords, time references, or field names directly signal the rule.\\
- \textbf{implicit}: The query requires applying the same rule, but the wording does NOT directly signal it. No obvious keywords --- the connection is indirect.\\
- \textbf{exception}: The query hits an edge case or exception path within the rule. It might seem like the rule's primary tool is correct at first glance, but the exception clause redirects to a different tool mentioned in the rule.\\[2pt]
You are generating "\{\{category\}\}" queries.\\
</CATEGORY>\\[4pt]
Generate exactly \{\{n\_queries\}\} diverse user queries where this business rule determines the correct tool selection. For each query, provide:\\
- The generated query\\
- The correct target tool in API/ENDPOINT format\\
- A one-sentence explanation of why the chosen tool is the target based on the rule\\[4pt]
Important context: This rule is FOR the tool "\{\{tool\_name\}\}". For explicit and implicit categories, the target tool should primarily be "\{\{tool\_name\}\}". For exception category, the target tool may be a DIFFERENT tool mentioned in the rule.\\[4pt]
Requirements:\\
1. Each query should be a realistic question a user would ask a system.\\
2. Use different parameters that could go as input to the query where applicable.\\
3. Vary phrasing: questions, imperatives, natural language, formal, informal\\
4. Do NOT reference the rule name, tool name, or any API/technical terminology in the query\\
5. Each query must have a different semantic focus --- avoid duplicates\\
6. The target tool MUST be one of the tools mentioned in the business rule
\end{tcolorbox}

\subsection{Reasoning Trace Generation}
\label{app:prompt-trace-gen}

The following prompt generates a structured reasoning trace for a given user query, as described in the trace generation step of \S\ref{subsec:stage2-data} 

\begin{tcolorbox}[colback=gray!3!white, colframe=gray!50!black, fonttitle=\small\bfseries, title={Prompt: Reasoning Trace Generator}, breakable]
\small\ttfamily
You are a tool selection reasoning engine. Given a user query, a set of candidate tools, and the correct tool(s) for the query, generate a structured reasoning trace that explains why the selected tool(s) are the best match.\\[4pt]
{}[START OF INPUT]\\
<USER QUERY> \{\{user\_query\}\} </USER QUERY>\\
<USER PERMISSION TARGETS> \{\{user\_query\_permission\_targets\}\} </USER PERMISSION TARGETS>\\
<CANDIDATE TOOLS> \{\{candidate\_pool\}\} </CANDIDATE TOOLS>\\
<SELECTED TOOLS> \{\{selected\_tools\}\} </SELECTED TOOLS>\\
{}[END OF INPUT]\\[4pt]
{}[START OF TOOL STRUCTURE]\\
Each tool has a tool\_name in the format API/ENDPOINT.\\
- API is the first part before the "/" --- it represents the API service.\\
- ENDPOINT is the second part after the "/" --- it represents a specific endpoint within that API service.\\
- Multiple tools can share the same API but differ in Endpoints.\\
{}[END OF TOOL STRUCTURE]\\[4pt]
{}[START OF TASK DESCRIPTION]\\
Generate a structured reasoning trace that explains why the selected tool(s) are the correct match for the user query. Follow the reasoning steps, and always follow business rule (if given below) that captures the discriminating principle for selection, and populate the output fields accordingly.\\
{}[END OF TASK DESCRIPTION]\\[4pt]
{}[START OF BUSINESS RULE]\\
\{\{business\_rule\}\} \\
{}[END OF BUSINESS RULE]\\[4pt]
{}[START OF REASONING STEPS]\\[2pt]
\#\# STEP 1: PERMISSION ACCESS FILTERING\\
- Each tool has permission targets representing required permissions.\\
- Exclude any tool whose permission targets share no overlap with the user's permission targets.\\
- Briefly state this filtering in 1-2 sentences. Do NOT name specific permission tags or tools. Simply describe the user's permission scope in general terms.\\[2pt]
\#\# STEP 2: SEMANTIC REASONING\\
Reason through why the selected tool(s) are the best match. Follow this hierarchy strictly:\\
1. API first: Identify and compare the most relevant services.\\
2. Then entities: Within the relevant API(s), explain why the selected endpoint(s) are correct over alternatives.\\
3. Conclude: Summarize why the selected tool(s) win.\\[2pt]
When candidate tools share the same API and differ only in endpoint, apply domain-specific routing rules to ground endpoint selection. When applying a routing principle, state the full rule text verbatim.\\[2pt]
Keep reasoning tight --- only discuss genuine contenders. Do not enumerate clearly irrelevant tools.\\
{}[END OF REASONING STEPS]\\[4pt]
{}[START OF OUTPUT FIELDS]\\
- permission\_filtering.reasoning: Step 1 access scope text.\\
- semantic\_reasoning.reasoning: Step 2 semantic reasoning text.\\
{}[END OF OUTPUT FIELDS]\\[4pt]
{}[START OF INSTRUCTIONS]\\
1. Only reason over tools explicitly provided --- do not hallucinate.\\
2. Use exact tool names as given --- no abbreviations.\\
3. Do not infer functionality beyond what is stated in descriptions.\\
4. Reason progressively: APIs -> Endpoints -> selection.\\
5. Permission step: do not mention specific Permission tags.\\
6. Trace must read as clean single-pass reasoning --- no mention of feedback.\\
7. No mention of candidate pool/set --- reason as if from own knowledge.\\
8. business\_rule must be followed if applicable and provided.\\
9. No meta-phrases referencing this prompt or instructions.\\
{}[END OF INSTRUCTIONS]
\end{tcolorbox}


\revision{
\section{Hybrid Retrieval Baseline}
\label{app:hybrid-baseline}

To establish whether \trace{}'s gains stem from the training curriculum rather than merely from access to business rules, we compare against a strong hybrid retrieval pipeline that receives the same rule supervision at inference time.

\paragraph{Pipeline.} \texttt{text-embedding-3-large} retrieves the top-50 candidates per query from the full tool catalog. An LLM reranker \texttt{claude-4.6-sonnet} then selects the suitable tool(s) from this candidate set. We test two variants: (i)~reranker receives only the query and candidate tool descriptions (\emph{no rules}), and (ii)~reranker additionally receives the business rules relevant to the candidate set (\emph{with rules}). The reranker is prompted to output a JSON list of selected tools, mirroring \trace{}'s output format.

\paragraph{Results.}

\begin{table}[H]
  \centering\tiny
  \begin{tabular}{lcc}
    \toprule
    Method & Domain~A R@gen & Domain~B R@gen \\
    \midrule
    Hybrid (no rules) & 67.2 & 54.5 \\
    Hybrid (with rules) & 66.4 & 56.1 \\
    \textbf{\trace{}} $(c, m, r{+}R)$ & \textbf{85.5} & \textbf{60.2} \\
    \bottomrule
  \end{tabular}
  \caption{Hybrid retrieval + LLM reranker vs.\ \trace{} on PRB.}
  \label{tab:hybrid-baseline}
\end{table}

\trace{} outperforms the hybrid pipeline by $+19.1$\,pp on Domain~A and $+4.1$\,pp on Domain~B (vs.\ the with-rules variant). Notably, injecting rules at inference time does not improve the hybrid pipeline, suggesting that the reranker struggles to apply rules over a large candidate set without parametric grounding. In contrast, \trace{}'s training-time internalization of rules enables effective disambiguation. Furthermore, \trace{} requires no external retrieval index or API calls at inference. The entire retrieval is a single forward pass through the model.
}

\revision{
\section{Human Labelling Study}
\label{app:human-study}

To provide external evidence that rule citation causally drives correct retrieval (rather than serving as post-hoc rationalization), we conducted a human labelling study on both PRB benchmarks.

\paragraph{Setup.} For the headline checkpoint \trace{} $(c, m, r{+}R)$, annotators examined all PRB queries whose ground-truth tool has an associated business rule. For each such query, the annotator checked whether the model's generated reasoning trace correctly cited the relevant business rule.

\paragraph{Results.}

\begin{table}[H]
  \centering\small
  \begin{tabular}{lcc}
    \toprule
    Benchmark & Rule cited (\%) & Correct retrieval (\%) \\
    \midrule
    PRB$_\text{A}$ & 82.4 & 71.2 \\
    PRB$_\text{B}$ & 79.5 & 75.3 \\
    \bottomrule
  \end{tabular}
  \caption{Human labelling study: when the ground-truth query requires a business rule, how often does \trace{} correctly cite it, and what is the corresponding retrieval accuracy.}
  \label{tab:human-study}
\end{table}

When the model correctly cites the governing rule, retrieval accuracy is substantially higher, consistent with the automated analysis in \S\ref{subsec:rule-grounding}. Combined with the monotonic scaling relationship in Figure~\ref{fig:pareto} (more rule data $\to$ higher recall) and the $94.6\%$ vs.\ $63.2\%$ recall gap between rule-citing and non-citing traces, this provides converging evidence that rule application is a causal driver of correct retrieval.
}

\revision{
\section{Extended Retrieval Metrics}
\label{app:extended-metrics}

Table~\ref{tab:extended-metrics} reports additional retrieval metrics for the headline checkpoint \trace{} $(c, m, r{+}R)$ under single-beam greedy decoding.

\begin{table}[H]
  \centering\small
  \begin{tabular}{lcc}
    \toprule
    Metric & Domain~A & Domain~B \\
    \midrule
    Avg $|\hat{A}(q)|$ & 2.63 & 2.01 \\
    Precision & 34.8 & 35.1 \\
    Recall (R@gen) & 85.5 & 60.2 \\
    F1 & 48.6 & 42.4 \\
    Off-vocabulary rate (\%) & 1.04 & 1.96 \\
    \bottomrule
  \end{tabular}
  \caption{Extended retrieval metrics for \trace{} $(c, m, r{+}R)$ under single-beam greedy decoding on PRB.}
  \label{tab:extended-metrics}
\end{table}

\paragraph{Interpreting precision.} Precision appears moderate (${\sim}35\%$) because the PRB gold labels annotate only a single valid tool-path per query, while multiple tool sequences can solve the user task in such large tool catalogs with near-duplicate tools. When the model retrieves 2--3 tools (avg $|\hat{A}(q)|{=}2.63$), some may be legitimate alternatives not annotated as gold. Precision is therefore a conservative lower bound in this setting, and recall is the more reliable diagnostic. The compact output set size (${\sim}2$--$3$ tools) is itself a practical advantage: compared to a fixed top-$K$ retrieval pipeline returning 10 candidates, \trace{}'s smaller set reduces downstream orchestrator confusion from semantically overlapping near-duplicates.

\paragraph{Off-vocabulary handling.} The off-vocabulary rate ($1.04\%$/$1.96\%$) measures how often the model generates a token string that does not correspond to any valid tool in $\mathcal{C}$. These hallucinated tokens are silently discarded at inference---they are never included in the retrieved set $\hat{A}(q)$. This rate is remarkably low given that the model selects from $8{,}283$ valid tokens without any trie-based constraint at decoding time.
}

\end{document}